\PassOptionsToPackage{table}{xcolor}
\documentclass[sigconf]{acmart}
\usepackage{subcaption}
\usepackage{makecell}
\usepackage{placeins}
\usepackage{multirow} 
\usepackage{algorithm}
\usepackage{algpseudocode}
\usepackage{cuted}

\setcopyright{none} 
\renewcommand\footnotetextcopyrightpermission[1]{} 

\definecolor{myorange}{RGB}{230,120,20}

\AtBeginDocument{%
  }

\acmConference[MMAsia]{ACM Multimedia Asia 2026}{2026}{VinUniversity, Hanoi, Vietnam}

\acmSubmissionID{441}

\begin{document}

\title{DAP-MR: Degradation-Aware Prompt–Magnitude Routing for Real-World Ultrasound Super-Resolution}

\author{Ngoc-Hoang-Lam Le}
\email{lengochoanglam.ee12@nycu.edu.tw}
\affiliation{%
  \institution{National Yang Ming Chiao Tung University}
  \city{Hsinchu}
  \country{Taiwan}
}

\author{Thi-An Le}
\email{anle.ee14@nycu.edu.tw}
\affiliation{%
  \institution{National Yang Ming Chiao Tung University}
  \city{Hsinchu}
  \country{Taiwan}
  }

\author{Dinh-Hung Le}
\email{ledinhhung332000.ee14@nycu.edu.tw}
\affiliation{%
  \institution{National Yang Ming Chiao Tung University}
  \city{Hsinchu}
  \country{Taiwan}
  }

\author{Manh-Hung Nguyen}
\email{hungnm@hcmute.edu.vn}
\affiliation{%
  \institution{Ho Chi Minh City University of Technology and Education}
  \city{Ho Chi Minh}
  \country{Vietnam}
  }

\author{Kai-Lung Hua}
\email{hua@mail.ntust.edu.tw}
\affiliation{%
  \institution{National Taiwan University of Science and Technology}
  \city{Taipei}
  \country{Taiwan}
  }

\author{Ching-Chun Huang}
\email{chingchun@nycu.edu.tw}
\correspondingauthor
\affiliation{%
  \institution{National Yang Ming Chiao Tung University}
  \city{Hsinchu}
  \country{Taiwan}
}

\renewcommand{\shortauthors}{Le et al.}

\begin{abstract}
When super-resolution models designed for natural images are used on ultrasound scans, relying on semantic text prompts can lead to unreliable results. The tags these models generate often do not match the actual restoration needs, and specialized medical terms may be misunderstood by general-purpose text encoders. Instead, prompts focused on common image degradations are more directly useful for guiding restoration. However, using a single prompt to describe both the type and severity of image degradation can entangle these factors, making it hard for the model to know how strong the degradation is. To address this limitation, we introduce Degradation-Aware Prompt--Magnitude Routing (DAP-MR). Each degradation type is represented by a dedicated token initialized from the contextualized EOS embedding of a degradation-oriented prompt and subsequently adapted during ultrasound SR training. A separate magnitude router predicts a continuous, input-dependent magnitude that modulates the corresponding token. Thus, language provides a degradation-aware initialization for \emph{what} should be restored, while magnitude routing supplies an explicit signal for \emph{how strongly} it should influence restoration. DAP-MR can be added to both one-step and multi-step diffusion super-resolution frameworks without changing their core structures. Our experiments—on synthetic datasets with both in-domain and under anatomical-domain shift, as well as unpaired real-world ultrasound data—show that DAP-MR delivers better restoration quality and generalizes more consistently across different domains.
\end{abstract}




\maketitle
\section{Introduction}
\label{ch:intro}


\begin{figure*}[t]
  \centering
  \includegraphics[
        width=0.93\textwidth,
        trim=0 0 0 0,
        clip
    ]{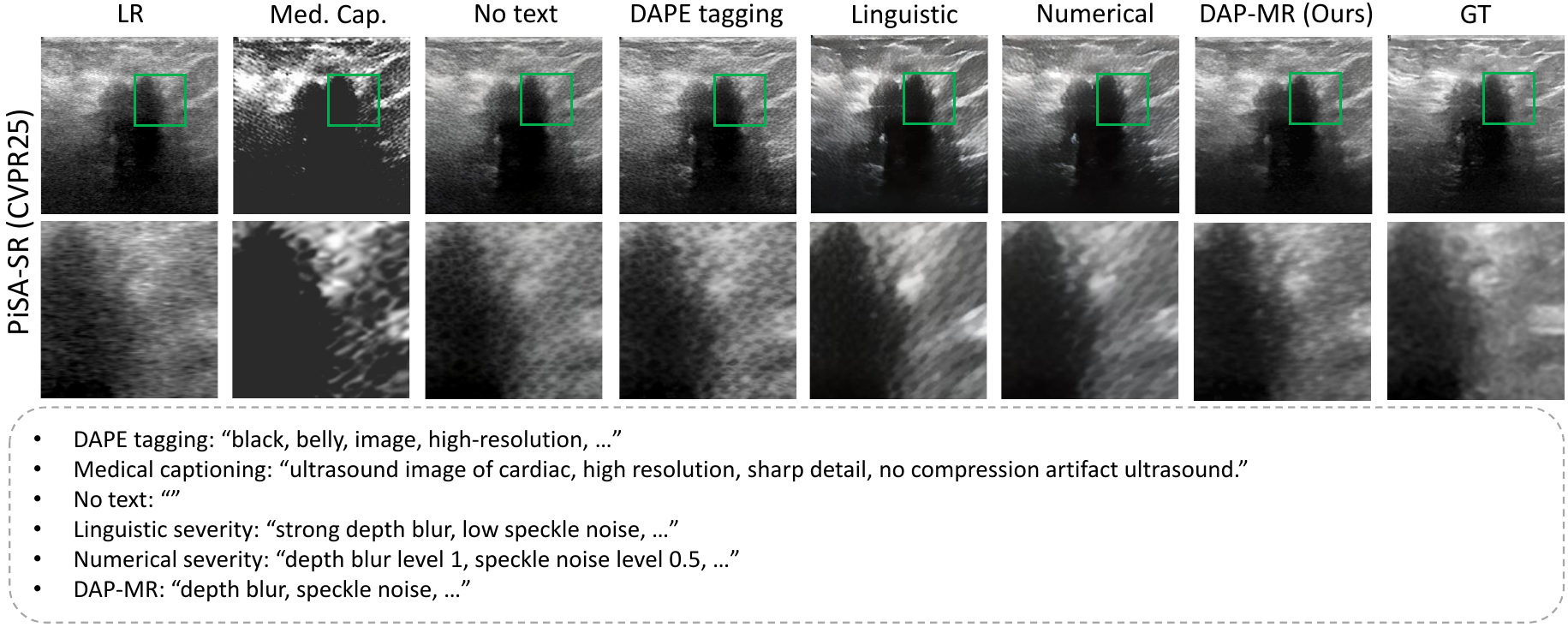}
  \vspace{-3pt}
  \caption{Motivating comparison of prompt strategies using the same PiSA-SR backbone. Natural tags produce limited changes over the no-text baseline, while medical captions may trigger unintended restoration behavior. Degradation prompts provide stronger restoration guidance, but may introduce over-smoothing or artifacts. DAP-MR separates degradation identity from magnitude, producing a result closer to the ground truth. The bottom row shows enlarged regions marked by the green boxes.}
  \label{fig:motivation}
\end{figure*}


Ultrasound super-resolution (SR) aims to recover fine structures from low-quality observations affected by limited spatial resolution and complex acquisition degradations. Compared with the extensive progress in natural-image SR, ultrasound-specific methods remain relatively limited and often rely on less flexible, task-specific architectures~\cite{ul_sr24,ul_sr25,ul_sr26}. In contrast, recent natural-image diffusion SR models provide stronger generative priors and more advanced conditioning mechanisms, motivating their use for ultrasound restoration. However, simply fine-tuning these models on ultrasound may not fully transfer their text-guided capabilities, as prompts designed for natural images can remain poorly aligned with ultrasound restoration.

Directly reusing prompt strategies designed for natural-image SR may provide limited guidance for ultrasound restoration. As shown in Fig. 1, the degradation-aware prompt extractor (DAPE), introduced in SeeSR \cite{seesr} and later adopted by PiSA-SR \cite{pisasr} and TVT \cite{tvt}, generates tags that mainly describe generic visual attributes rather than the ultrasound structures or degradations to be restored. They therefore produce only limited changes compared with the no-text baseline. Medical captions provide less predictable restoration guidance in our comparison, suggesting that domain-specific descriptions are not always effective when directly transferred to general-purpose text-conditioned SR models.

Degradation-oriented prompts reduce the dependence on domain-specific terminology by describing common image defects, such as blur or noise. As shown in Fig.~\ref{fig:motivation}, they produce more noticeable restoration changes, indicating that degradation information is more directly relevant to SR. However, existing text-based methods encode both degradation identity and severity within the same prompt, using expressions such as ``strong blur''~\cite{promptsr} or numerical descriptions such as ``blur level 0.8''~\cite{spire}. Although these prompts provide useful restoration semantics, both factors must be conveyed through the same text-conditioning representation. Their effects may become entangled, resulting in ambiguous strength control and potentially causing over-smoothing or artifacts.

To address this limitation, we propose Degradation-Aware Prompt--Magnitude Routing (DAP-MR), a factorized conditioning framework that separates degradation identity from degradation magnitude. Each degradation is represented by a token initialized from the EOS representation of a degradation-oriented CLIP prompt and subsequently adapted during ultrasound SR training. A magnitude router predicts a continuous, input-dependent magnitude to modulate the corresponding token. Thus, language provides a degradation-specific semantic initialization, while magnitude routing separately controls the strength of its contribution. However, when all training samples are supervised only toward the clean HR target, the SR backbone has limited incentive to rely on the degradation-token conditioning, since restoration can be learned largely from visual features and the backbone prior. We therefore introduce two complementary supervision strategies. Partial-dose supervision associates different control magnitudes with intermediate restoration targets, while mismatch-specificity supervision discourages responses to non-matching degradation tokens. Together, they encourage the model to utilize both degradation identity and magnitude during restoration.

\noindent\textbf{Our contributions are summarized as follows:}
\begin{itemize}
    \setlength{\topsep}{2pt}
    \setlength{\itemsep}{1pt}
    \setlength{\parsep}{0pt}
    \setlength{\parskip}{0pt}

     \item We propose DAP-MR, which represents degradation identity using CLIP-initialized, ultrasound-adapted EOS tokens and represents degradation strength using continuous input-dependent magnitudes predicted by a router.

    \item We introduce a training strategy that encourages the SR model to better utilize both magnitude and degradation-specific conditioning.

    \item Experiments across one-step and multi-step diffusion SR backbones demonstrate improved restoration quality and generalization to anatomy-shifted and real-world ultrasound data.
\end{itemize}
\vspace{-5pt}
\section{Related Work}
\label{ch:related_works}

Recent diffusion-based super-resolution (SR) methods leverage the generative priors and conditioning pathways of pre-trained text-to-image models. SeeSR~\cite{seesr} extracts semantic tags from LR inputs, while PiSA-SR~\cite{pisasr} and TVT~\cite{tvt} improve the underlying diffusion SR backbone. However, transferring semantic prompting to ultrasound remains difficult: automatically generated tags may be only loosely related to the ultrasound image, while medical captions rely on specialized terminology that may not be reliably interpreted by general-purpose text encoders. Degradation-oriented prompts offer a more transferable alternative because they are less dependent on domain-specific semantics and describe restoration-relevant image defects directly. PromptSR~\cite{promptsr} and DaLPSR~\cite{dalpsr} represent degradation characteristics through language, whereas S3Diff~\cite{s3diff} predicts degradation-aware features for restoration modulation. Nevertheless, these methods either encode degradation identity and severity within the same text-conditioning stream or lack explicit continuous control over the magnitude of each degradation.

Ultrasound SR methods have explored adversarial learning~\cite{cyclegan}, progressive residual reconstruction~\cite{prlmu}, and diffusion-based restoration~\cite{ul_sr24}. Many existing approaches learn a fixed restoration mapping from paired or synthetically degraded LR--HR data. In real-world ultrasound, however, degradation types and strengths vary across scanners and acquisition conditions and may not be fully captured by synthetic pipelines. This motivates an input-adaptive framework that controls each degradation separately, together with evaluation on unpaired real-world ultrasound data.

\vspace{-4pt}
\section{Proposed Method}
\label{ch:method}

\begin{figure*}[t]
    \centering
    \includegraphics[width=0.9\textwidth]{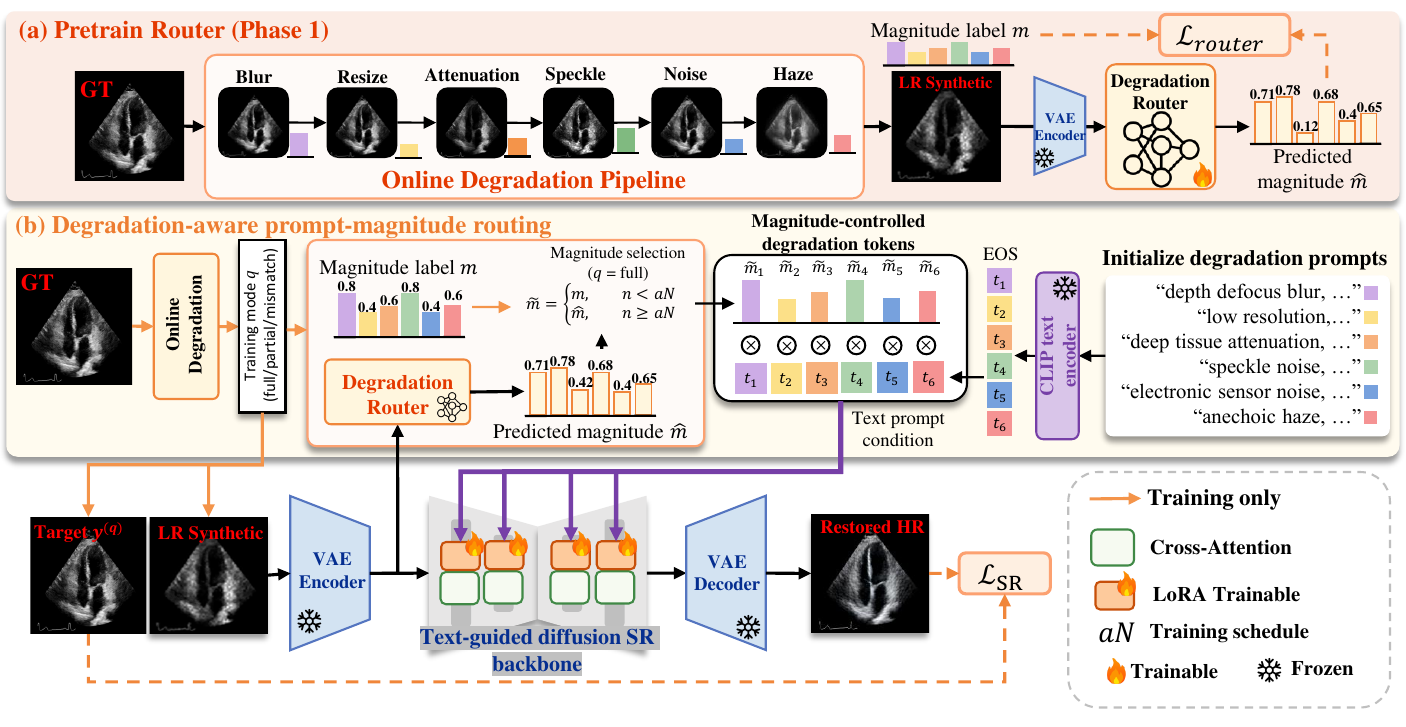} 
    \caption{
\textbf{Overview of DAP-MR.}
(a) The magnitude router is pretrained on synthetic LR images with continuous degradation-magnitude labels. (b) Degradation tokens are initialized from contextualized CLIP EOS representations and jointly optimized with the SR backbone. A training mode $q\in\{\textit{full},\textit{partial},\textit{mismatch}\}$ determines the target $\mathbf{y}^{(q)}$ and control magnitude $\tilde{\mathbf{m}}^{(q)}$. Full, partial-dose, and mismatch-specificity supervision learn the clean endpoint, magnitude-dependent restoration, and degradation-selective responses, respectively.
The resulting magnitude-modulated degradation tokens condition the original text-guided SR pathway.
}
    \label{fig:architecture}
\end{figure*}

\subsection{Problem Formulation}

Let $\mathbf{x}_{H}$ denote a clean high-resolution ultrasound image and $\mathbf{x}_{L}$ its degraded observation. In blind SR, $\mathbf{x}_{L}$ may contain any subset of $K$ degradation categories with unknown magnitudes:
\begin{equation}
    \mathbf{x}_{L}
    =
    \mathcal{G}(\mathbf{x}_{H};\mathbf{m}),
    \qquad
    \mathbf{m}
    =
    [m_1,\ldots,m_K]\in[0,1]^K,
    \label{eq:degradation}
\end{equation}
where $\mathcal{G}$ denotes the degradation process, $m_k=0$ indicates that degradation $k$ is absent, and a larger $m_k$ indicates stronger degradation. Many existing text-based degradation prompting methods encode degradation identity and severity jointly in language. DAP-MR instead represents degradation identity through text-derived tokens and modulates their conditioning using continuous magnitudes.

Given the LR input and its prompt--magnitude condition $\mathbf{P}$, the SR backbone produces
\vspace{-3pt}
\begin{equation}
    \hat{\mathbf{x}}_{H}
    =
    F_{\theta}(\mathbf{x}_{L},\mathbf{P}).
    \label{eq:sr_formulation}
\end{equation}
The construction of $\mathbf{P}$ is detailed in Sec.~\ref{sec:prompt_magnitude_routing}.
\vspace{-2pt}
\subsection{Router Pretraining}
\label{sec:online_degradation}
Training the SR model using only ground-truth magnitudes creates a train--inference gap, because inference relies on router-predicted magnitudes that may deviate from the exact labels seen during training. However, using predictions from an uninitialized router at the beginning of SR training may provide inaccurate control signals and hinder restoration learning. We therefore pretrain the magnitude router on synthetically degraded ultrasound images with known continuous magnitude labels before integrating it into SR training.

\textbf{Online ultrasound degradation synthesis.} We generate training pairs online using a degradation pipeline that combines generic real-world degradations adopted from Real-ESRGAN~\cite{realesrgan} with ultrasound-specific effects. As shown in Fig.~\ref{fig:architecture}, we consider $K=6$ degradation categories. Each degradation $k$ is controlled by a degradation parameter $p_k$ sampled from a predefined plausible range $[p_k^{\min},p_k^{\max}]$. Because these parameters have different scales and meanings, we map them to a unified magnitude range:
\begin{equation}
    m_k=\psi_k(p_k)\in[0,1],
\end{equation}
where $\psi_k$ is defined such that a larger $m_k$ consistently indicates a stronger degradation. An inactive degradation is assigned $m_k=0$. The parameter ranges are restricted to avoid unrealistically destructive cases that remove useful anatomical information. At each iteration, one or multiple degradations and their magnitudes are randomly sampled and applied through the degradation process in Eq.~\ref{eq:degradation}, producing a synthetic LR image together with its exact magnitude vector.

\textbf{Router optimization.} Rather than only classifying whether each degradation is present, the router performs multi-dimensional regression to estimate its magnitude. The synthesized LR image is first encoded by the frozen VAE encoder $E$, and the resulting latent representation is passed to the router $R_{\phi}$:
\begin{equation}
    \mathbf{z}=E(\mathbf{x}_{L}),
    \qquad
    \hat{\mathbf{m}}
    =
    \sigma\!\left(R_{\phi}(\mathbf{z})\right),
    \qquad
    \hat{\mathbf{m}}\in[0,1]^K.
\end{equation}
The router is optimized using a Smooth-$L_1$ regression loss:
\begin{equation}
    \mathcal{L}_{\mathrm{router}}
    =
    \operatorname{SmoothL1}
    \left(
        \hat{\mathbf{m}},
        \mathbf{m}
    \right).
\end{equation}
After pretraining, the router estimates magnitudes of LR images that are subsequently introduced into SR training to reduce the gap between ground-truth conditioning and blind inference.
\vspace{-2pt}
\subsection{Degradation-Aware Prompt--Magnitude Routing}
\label{sec:prompt_magnitude_routing}

\noindent\textbf{Degradation prompt encoding.}
As illustrated in Fig.~\ref{fig:architecture}(b),we define a fixed degradation-oriented prompt $P_k$ for each of the $K$ degradation categories. Each prompt is encoded by the frozen CLIP text encoder $\mathcal{T}$, and its contextualized EOS representation is used to initialize a degradation token:
\begin{equation}
    \mathbf{t}_k^{(0)}=\mathcal{T}(P_k)_{\mathrm{EOS}}.
\end{equation}
The degradation tokens $\{\mathbf{t}_k\}_{k=1}^{K}$ are jointly optimized with the SR backbone during ultrasound training. This retains degradation-aware semantics from the language space while allowing the token representation to adapt to ultrasound appearance statistics.

\noindent\textbf{Magnitude routing.}
Rather than expressing degradation severity through linguistic or numerical text, DAP-MR applies a continuous magnitude directly to each degradation token:
\begin{equation}
    \mathbf{c}_k^{(n)}
    =
    \tilde{m}_k^{(n)}\mathbf{t}_k,
    \qquad
    \mathbf{P}^{(n)}
    =
    [\mathbf{c}_1^{(n)},\ldots,\mathbf{c}_K^{(n)}],
\end{equation}
where $\tilde{m}_k^{(n)}$ is the selected control magnitude at training iteration $n$. The adapted token $\mathbf{t}_k$ specifies the degradation identity, while $\tilde m_k^{(n)}$ modulates its contribution to the conditioning sequence. The resulting token sequence is passed through the original text-conditioning pathway of the SR backbone.

For the \textit{full-restoration} mode, the magnitude source switches from ground-truth labels to router predictions during training:
\begin{equation}
    \tilde{\mathbf{m}}^{(n)}
    =
    \begin{cases}
        \mathbf{m}, & n<aN,\\
        \hat{\mathbf{m}}, & n\geq aN,
    \end{cases}
\end{equation}
where $N$ is the total number of training iterations and $a\in[0,1]$ defines the switching ratio. Ground-truth magnitudes are used initially to establish the relationship between magnitude and restoration strength. Router predictions are then introduced to expose the SR model to the estimated conditions used during blind inference. The \textit{partial-dose} and \textit{mismatch-specificity} modes instead use their explicitly constructed magnitude vectors, as described in Sec.~\ref{sec:training_strategy}.

\subsection{Training Strategy}
\label{sec:training_strategy}

Standard SR training maps a degraded input directly to its clean HR target, which we refer to as full-restoration supervision. This mode establishes the clean restoration endpoint, but alone it does not define how the output should vary at intermediate control magnitudes or how the model should respond to a non-matching degradation token. We therefore use three complementary training modes: full restoration learns the complete restoration endpoint, partial-dose supervision provides intermediate targets for magnitude-dependent restoration, and mismatch-specificity supervision encourages each degradation token to affect only its corresponding degradation. The complete training procedure is provided in Supplementary Algorithm~2.

\noindent\textbf{Full-restoration supervision.}
In the full mode, one or multiple degradations are applied to an HR image $\mathbf{x}_{H}$, while the clean image is used as the target:
\begin{equation}
    \mathbf{y}^{(\mathrm{full})}
    =
    \mathbf{x}_{H}.
\end{equation}
The magnitude condition follows Sec.~\ref{sec:prompt_magnitude_routing}, using ground-truth magnitudes initially and router predictions near the end of training.

For the two additional modes, we synthesize an input containing a single degradation $k$:
\begin{equation}
    \mathbf{x}_{L}^{(k)}
    =
    \mathcal{G}_{k}(\mathbf{x}_{H};m_k).
    \label{eq:single_degradation}
\end{equation}
Using a single degradation in these modes isolates the response of one control dimension and avoids ambiguity caused by interactions among multiple degradation operators.

\noindent\textbf{Partial-dose supervision.}
We sample $u\in(0,1)$ and use only a fraction of the input degradation magnitude as the restoration control:
\begin{equation}
    \tilde{m}_k^{(\mathrm{partial})}
    =
    u m_k,
    \qquad
    \tilde{m}_l^{(\mathrm{partial})}
    =
    0
    \quad \forall l\neq k.
\end{equation}
The corresponding target retains the remaining degradation:
\begin{equation}
    \mathbf{y}^{(\mathrm{partial})}
    =
    \mathcal{G}_{k}
    \left(
        \mathbf{x}_{H};
        (1-u)m_k
    \right).
\end{equation}
This provides an intermediate target that explicitly relates the supplied control magnitude to the intended restoration strength.

\noindent\textbf{Mismatch-specificity supervision.}
We select a non-matching token $j\neq k$ and activate it using the magnitude of the degradation present in the input:
\begin{equation}
    \tilde{m}_{j}^{(\mathrm{mismatch})}
    =
    m_k,
    \qquad
    \tilde{m}_{l}^{(\mathrm{mismatch})}
    =
    0
    \quad \forall l\neq j.
\end{equation}
Because degradation $j$ is absent, activating its token should not alter degradation $k$. We therefore use the degraded input itself as the target:
\begin{equation}
    \mathbf{y}^{(\mathrm{mismatch})}
    =
    \mathbf{x}_{L}^{(k)}.
\end{equation}
This supervision encourages degradation-specific responses and suppresses cross-degradation effects.

Let $\mathbf{x}_{L}^{(q)}$, $\mathbf{y}^{(q)}$, and $\tilde{\mathbf{m}}^{(q)}$ denote the input, target, and control-magnitude vector constructed for training mode $q$. The corresponding prompt condition $\mathbf{P}^{(q)}$ is constructed following Sec.~\ref{sec:prompt_magnitude_routing}. The SR model is optimized using
\begin{equation}
    \mathcal{L}_{\mathrm{SR}}^{(q)}
    =
    \mathcal{L}_{\mathrm{base}}
    \left(
        \theta;
        \mathbf{x}_{L}^{(q)},
        \mathbf{y}^{(q)},
        \mathbf{P}^{(q)}
    \right),
\end{equation}
where $\mathcal{L}_{\mathrm{base}}$ denotes the original backbone-specific training objective. Depending on the backbone, it may include diffusion denoising, reconstruction, and perceptual terms. We retain the original loss terms, weights, and optimization settings of each backbone. DAP-MR therefore modifies the conditioning and supervision targets without introducing an additional restoration loss.
\definecolor{gainbg}{RGB}{225,239,255}
\definecolor{lossbg}{RGB}{255,232,232}
\definecolor{gaintext}{RGB}{0,78,150}
\definecolor{losstext}{RGB}{170,35,35}

\newcommand{\gain}[2]{%
    \cellcolor{gainbg}%
    \mbox{#1\,\textcolor{gaintext}{\scriptsize $(+#2\%)$}}%
}

\newcommand{\loss}[2]{%
    \cellcolor{lossbg}%
    \mbox{#1\,\textcolor{losstext}{\scriptsize $(-#2\%)$}}%
}

\section{Experimental Results}
\label{ch:experiments}
\subsection{Experimental Setup}


We extract 1,417 echocardiography frames from EchoNet~\cite{stanford2020echonet}, using an 8:2 patient-level train--test split. Paired LR--HR data are synthesized using the degradation pipeline in Sec.~\ref{sec:online_degradation}, with the magnitude router pretrained exclusively on synthetic Cardiac training data and corresponding magnitude labels. We integrate DAP-MR into two multi-step backbones, SeeSR~\cite{seesr} and StableSR~\cite{stablesr}, and two one-step backbones, PiSA-SR~\cite{pisasr} and TVT~\cite{tvt}. All models are fine-tuned for up to $8.5$K iterations on $512\times512$ patches, with each baseline and its DAP-MR variant sharing identical data, training configurations, and backbone-specific losses.

For paired synthetic evaluation, we use Cardiac as the in-domain benchmark and BUSI~\cite{busi} (768 breast ultrasound images) to assess anatomical-domain generalization. Both use the same synthetic degradation pipeline; thus, BUSI represents an anatomical shift rather than unseen degradations. For unpaired real-world evaluation, we use CAMUS-Poor~\cite{camus} (232 low-quality echocardiography images) and HMC-QU~\cite{HMC_QU} (473 clinical ultrasound images), neither used in training nor synthetically degraded. We evaluate paired reconstruction using PSNR, SSIM~\cite{psnr}, and LPIPS~\cite{lpips}, and assess unpaired real-world performance using Dice scores from the frozen pretrained SAMUS segmentation model~\cite{SAMUS}, as HR references are unavailable.



\begin{table}[t]
    \centering
    \caption{
    Cross-backbone comparison on the paired synthetic Cardiac and BUSI test
    sets. Cardiac and BUSI represent in-domain and out-of-domain evaluation,
    respectively. Percentages shown beside each DAP-MR result indicate relative improvement (positive indicates better performance) from the corresponding baseline.
    }
    \label{tab:cross_backbone_sr}

    \scriptsize
    \setlength{\tabcolsep}{4pt}
    \renewcommand{\arraystretch}{1.15}

    \resizebox{\columnwidth}{!}{%
    \begin{tabular}{llcccc}
        \toprule
        \textbf{Backbone}
        & \textbf{Variant}
        & \textbf{Steps}
        & \textbf{PSNR}$\uparrow$
        & \textbf{SSIM}$\uparrow$
        & \textbf{LPIPS}$\downarrow$ \\
        \midrule

        \multicolumn{6}{l}{\textbf{Cardiac: In-domain}} \\
        \midrule

        \multirow{2}{*}{SeeSR}
        & Baseline
        & \multirow{2}{*}{50}
        & $26.815$ & $0.8888$ & $0.0718$ \\
        & + DAP-MR
        &
        & \loss{$26.723$}{0.34}
        & \gain{$0.8901$}{0.15}
        & \loss{$0.0748$}{4.18} \\
        
        \addlinespace[2pt]
        
        \multirow{2}{*}{StableSR}
        & Baseline
        & \multirow{2}{*}{50}
        & $26.650$ & $0.8873$ & $0.0813$ \\
        & + DAP-MR
        &
        & \gain{$26.678$}{0.11}
        & \gain{$0.8881$}{0.09}
        & \loss{$0.0822$}{1.11} \\
        
        \addlinespace[2pt]
        
        \multirow{2}{*}{TVT}
        & Baseline
        & \multirow{2}{*}{1}
        & $26.523$ & $0.9038$ & $0.0724$ \\
        & + DAP-MR
        &
        & \gain{$27.104$}{2.19}
        & \gain{$0.9077$}{0.43}
        & \gain{$0.0693$}{4.28} \\
        
        \addlinespace[2pt]
        
        \multirow{2}{*}{PiSA-SR}
        & Baseline
        & \multirow{2}{*}{1}
        & $26.640$ & $0.8870$ & $0.0802$ \\
        & + DAP-MR
        &
        & \gain{$26.881$}{0.90}
        & \gain{$0.8943$}{0.82}
        & \loss{$0.0842$}{4.99} \\
        
        \midrule
        \multicolumn{6}{l}{\textbf{BUSI: Out-of-domain}} \\
        \midrule
        
        \multirow{2}{*}{SeeSR}
        & Baseline
        & \multirow{2}{*}{50}
        & $24.151$ & $0.6819$ & $0.4751$ \\
        & + DAP-MR
        &
        & \gain{$24.773$}{2.58}
        & \gain{$0.7185$}{5.37}
        & \gain{$0.4321$}{9.05} \\
        
        \addlinespace[2pt]
        
        \multirow{2}{*}{StableSR}
        & Baseline
        & \multirow{2}{*}{50}
        & $23.796$ & $0.6305$ & $0.4652$ \\
        & + DAP-MR
        &
        & \gain{$23.900$}{0.44}
        & \gain{$0.6311$}{0.10}
        & \loss{$0.4744$}{1.98} \\
        
        \addlinespace[2pt]
        
        \multirow{2}{*}{TVT}
        & Baseline
        & \multirow{2}{*}{1}
        & $23.984$ & $0.6360$ & $0.4616$ \\
        & + DAP-MR
        &
        & \gain{$24.220$}{0.98}
        & \gain{$0.7083$}{11.37}
        & \gain{$0.4451$}{3.57} \\
        
        \addlinespace[2pt]
        
        \multirow{2}{*}{PiSA-SR}
        & Baseline
        & \multirow{2}{*}{1}
        & $23.504$ & $0.6074$ & $0.3904$ \\
        & + DAP-MR
        &
        & \gain{$25.045$}{6.56}
        & \gain{$0.7616$}{25.39}
        & \gain{$0.3460$}{11.37} \\

        \bottomrule
    \end{tabular}%
    }
\end{table}

\subsection{Quantitative Results}
\label{sec:quantitative_results}

Table~\ref{tab:cross_backbone_sr} evaluates DAP-MR across four diffusion SR backbones. Overall, DAP-MR improves 15 of the 16 PSNR and SSIM comparisons. On the in-domain Cardiac set, SSIM improves for all four backbones, while PSNR improves for StableSR, TVT, and PiSA-SR and decreases slightly for SeeSR by $0.092$ dB. TVT obtains the largest Cardiac PSNR gain of $+0.581$ dB, while PiSA-SR improves from $26.640$ to $26.881$ dB. On the anatomy-domain-shifted BUSI set, DAP-MR improves both PSNR and SSIM for all four backbones. PiSA-SR shows the largest gains, increasing PSNR from $23.504$ to $25.045$ dB ($+1.541$ dB) and SSIM from $0.6074$ to $0.7616$ ($+0.1542$). LPIPS improves in four of the eight backbone--dataset comparisons. These results indicate that DAP-MR provides relatively modest and occasionally mixed improvements in-domain, while its benefits are more pronounced and consistent under anatomical-domain shift. LPIPS improves in four of the eight backbone--dataset comparisons, potentially reflecting backbone-dependent trade-offs and the limited transferability of natural-image-pretrained perceptual features to ultrasound.

Table~\ref{tab:downstream_segmentation} evaluates downstream segmentation on the two unpaired real-world datasets. DAP-MR increases Dice scores in all eight backbone--dataset comparisons, although the magnitude of improvement varies across backbones. TVT shows the largest gains, improving by $3.10$ Dice points on CAMUS-Poor and $9.04$ points on HMC-QU. PiSA-SR shows a negligible change on CAMUS-Poor ($74.23$ to $74.26$, $+0.03$ points) but improves substantially on HMC-QU from $52.94$ to $57.32$ ($+4.38$ points). These results suggest that DAP-MR can improve compatibility with the frozen downstream segmentation model on real ultrasound images, while the small gains observed for some backbone--dataset pairs should not be overinterpreted. 

\begin{table}[t!]
    \centering
    \caption{
        \textcolor{myorange}{\textbf{Downstream segmentation}} performance using the frozen SAMUS evaluator.
        Dice scores are reported in percentage.
        Values in parentheses indicate relative changes from the corresponding baseline.
    }
    \label{tab:downstream_segmentation}

    \scriptsize
    \setlength{\tabcolsep}{4pt}
    \renewcommand{\arraystretch}{1.12}

    \begin{tabular}{llcc}
    \toprule
    \textbf{Backbone}
    & \textbf{Variant}
    & \textbf{CAMUS-Poor}$\uparrow$
    & \textbf{HMC-QU}$\uparrow$ \\
    \midrule

    \multirow{2}{*}{SeeSR}
    & Baseline
    & $75.00$
    & $59.07$ \\
    & + DAP-MR
    & \gain{$75.30$}{0.40}
    & \gain{$59.66$}{1.00} \\

    \addlinespace[2pt]

    \multirow{2}{*}{StableSR}
    & Baseline
    & $74.27$
    & $59.97$ \\
    & + DAP-MR
    & \gain{$74.71$}{0.59}
    & \gain{$60.07$}{0.17} \\

    \addlinespace[2pt]

    \multirow{2}{*}{TVT}
    & Baseline
    & $71.70$
    & $48.15$ \\
    & + DAP-MR
    & \gain{$74.80$}{4.32}
    & \gain{$57.19$}{18.77} \\

    \addlinespace[2pt]

    \multirow{2}{*}{PiSA-SR}
    & Baseline
    & $74.23$
    & $52.94$ \\
    & + DAP-MR
    & \gain{$74.26$}{0.04}
    & \gain{$57.32$}{8.27} \\

    \bottomrule
    \end{tabular}
\end{table}

\subsection{Qualitative Results}
\begin{figure*}[t]
    \centering
    \includegraphics[width=0.7\textwidth]{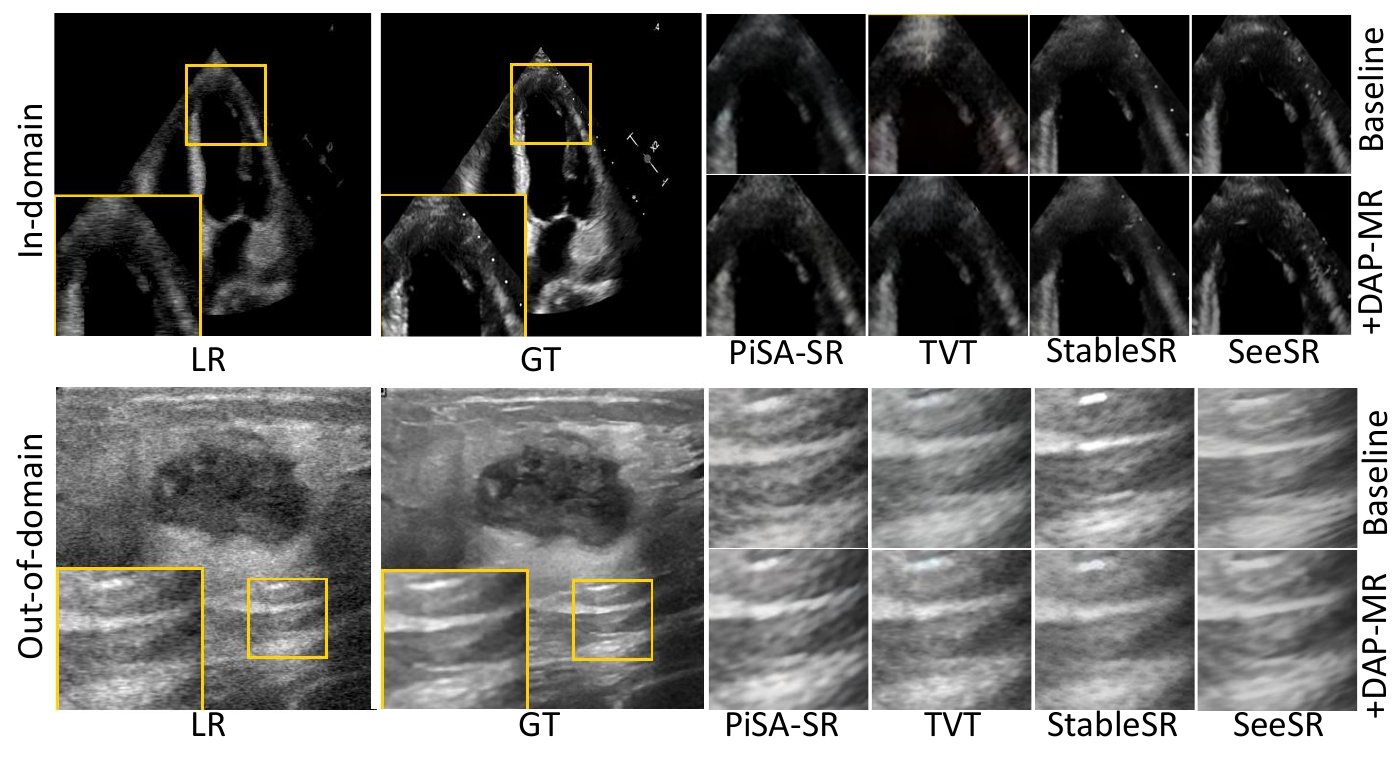} 
    \caption{Qualitative comparison of baseline and DAP-MR results on the in-domain Cardiac and out-of-domain BUSI test sets across PiSA-SR, TVT, StableSR, and SeeSR. Yellow boxes indicate regions of interest for local comparison.
    }
    \label{fig:qualitative}
\end{figure*}

Figure~\ref{fig:qualitative} compares the baseline and DAP-MR outputs across four diffusion SR backbones. On the in-domain Cardiac set, the visual differences are generally moderate, consistent with the relatively small quantitative gains. PiSA-SR and TVT show clearer local boundaries and reduced residual blur in the highlighted regions, whereas StableSR and SeeSR exhibit smaller changes. On the anatomy-domain-shifted BUSI set, DAP-MR produces more noticeable improvements in local structure and ultrasound texture without target-domain fine-tuning. Importantly, the benefit is not uniform across backbones, which is also reflected in the reconstruction metrics. These examples suggest that DAP-MR is most beneficial when the original text-conditioning pathway provides insufficient degradation-aware guidance, particularly under anatomical-domain shift.

\subsection{Ablation Study}

Table~\ref{tab:prompt_initialize} evaluates degradation-token representation, supervision strategy, and EOS-token adaptation. Replacing the full degradation text with fixed EOS tokens substantially improves BUSI performance, suggesting that a compact degradation-specific representation transfers better than directly encoding the complete prompt. Partial-dose and mismatch-specificity supervision provide complementary gains by improving magnitude-aware restoration and encouraging degradation-selective responses. Allowing the CLIP-initialized EOS tokens to remain learnable during SR training further improves cross-domain performance, with the final configuration reaching $26.881$ dB PSNR and $0.8943$ SSIM on Cardiac, $25.045$ dB PSNR and $0.7616$ SSIM on BUSI, and $57.32$ Dice on HMC-QU. These results suggest that CLIP provides a useful degradation-semantic initialization, while ultrasound-domain adaptation improves transfer beyond the training anatomy.

Table~\ref{tab:magnitude_selection} studies the magnitude source used during full-restoration training. Router-only conditioning performs worse on most reconstruction metrics, indicating that predicted magnitudes are less reliable when used throughout training, whereas ground-truth magnitudes provide more stable supervision. The scheduled ground-truth-to-router strategy gradually exposes the SR backbone to the estimated conditions encountered during blind inference, providing a practical compromise between accurate synthetic supervision and inference-time robustness. We therefore adopt this scheduled transition together with the learnable-EOS representation in the final DAP-MR configuration.

\begin{table}[t]
\centering
\caption{
Ablation of degradation-token representation and supervision strategies using PiSA-SR.
Cardiac is in-domain, while BUSI, CAMUS-Poor, and HMC-QU are unseen during training.
Partial-only and mismatch-only variants use a 40\% sampling ratio.
Best and second-best results are shown in bold and underlined.
}
\label{tab:prompt_initialize}

\scriptsize
\setlength{\tabcolsep}{1.3pt}
\renewcommand{\arraystretch}{1.10}

\resizebox{\columnwidth}{!}{%
\begin{tabular}{ccccccccccccc}
\toprule
\multirow{2}{*}{\shortstack{\textbf{Full}\\\textbf{token}}}
&
\multirow{2}{*}{\shortstack{\textbf{EOS}\\\textbf{token}}}
&
\multirow{2}{*}{\shortstack{\textbf{Token}\\\textbf{update}}}
&
\multirow{2}{*}{\textbf{Partial}}
&
\multirow{2}{*}{\textbf{Mismatch}}
&
\multicolumn{3}{c}{\textbf{Cardiac}}
&
\multicolumn{3}{c}{\textbf{BUSI}}
&
\multirow{2}{*}{\shortstack{\textbf{CAMUS-P}\\\textbf{Dice}$\uparrow$}}
&
\multirow{2}{*}{\shortstack{\textbf{HMC-QU}\\\textbf{Dice}$\uparrow$}}
\\

\cmidrule(lr){6-8}
\cmidrule(lr){9-11}

&
&
&
&
&
\textbf{PSNR}$\uparrow$
&
\textbf{SSIM}$\uparrow$
&
\textbf{LPIPS}$\downarrow$
&
\textbf{PSNR}$\uparrow$
&
\textbf{SSIM}$\uparrow$
&
\textbf{LPIPS}$\downarrow$
&
&
\\

\midrule

$\checkmark$
&
&
--
&
&
&
$26.394$
& $0.8866$
& $\mathbf{0.0750}$
& $22.885$
& $0.6814$
& $0.4291$
& $73.83$
& $52.46$
\\

&
$\checkmark$
&
Fixed
&
&
&
$26.550$
& $0.8854$
& $0.0777$
& $24.112$
& $\underline{0.7223}$
& $0.3896$
& $73.84$
& $54.32$
\\

&
$\checkmark$
&
Fixed
&
$\checkmark$
&
&
$26.425$
& $0.8842$
& $0.0797$
& $24.391$
& $0.7069$
& $0.3775$
& $74.11$
& $\underline{56.10}$
\\

&
$\checkmark$
&
Fixed
&
&
$\checkmark$
&
$26.426$
& $0.8835$
& $0.0819$
& $\underline{24.618}$
& $0.7211$
& $\underline{0.3462}$
& $73.92$
& $53.29$
\\

&
$\checkmark$
&
Fixed
&
$\checkmark$
&
$\checkmark$
&
$\underline{26.717}$
& $\underline{0.8900}$
& $\underline{0.0769}$
& $24.479$
& $0.7119$
& $0.3796$
& $\mathbf{74.26}$
& $56.08$
\\

&
$\checkmark$
&
Learnable
&
$\checkmark$
&
$\checkmark$
&
$\mathbf{26.881}$
& $\mathbf{0.8943}$
& $0.0842$
& $\mathbf{25.045}$
& $\mathbf{0.7616}$
& $\mathbf{0.3460}$
& $\mathbf{74.26}$
& $\mathbf{57.32}$
\\

\bottomrule
\end{tabular}%
}
\end{table}

\begin{table}[t]
\centering
\caption{
Ablation of magnitude sources used during full-restoration training.
``Label'' uses ground-truth degradation magnitudes, ``Router'' uses router-predicted magnitudes, and enabling both denotes the scheduled transition from label to router magnitudes.
Best and second-best results are shown in bold and underlined.
}
\label{tab:magnitude_selection}

\scriptsize
\setlength{\tabcolsep}{1.8pt}
\renewcommand{\arraystretch}{1.10}

\resizebox{\columnwidth}{!}{%
\begin{tabular}{cccccccccc}
\toprule
\multirow{2}{*}{\shortstack{\textbf{Label}\\\textbf{mag.}}}
&
\multirow{2}{*}{\shortstack{\textbf{Router}\\\textbf{mag.}}}
&
\multicolumn{3}{c}{\textbf{Cardiac}}
&
\multicolumn{3}{c}{\textbf{BUSI}}
&
\multirow{2}{*}{\shortstack{\textbf{CAMUS-P}\\\textbf{Dice}$\uparrow$}}
&
\multirow{2}{*}{\shortstack{\textbf{HMC-QU}\\\textbf{Dice}$\uparrow$}} \\

\cmidrule(lr){3-5}
\cmidrule(lr){6-8}

&
&
\textbf{PSNR}$\uparrow$
&
\textbf{SSIM}$\uparrow$
&
\textbf{LPIPS}$\downarrow$
&
\textbf{PSNR}$\uparrow$
&
\textbf{SSIM}$\uparrow$
&
\textbf{LPIPS}$\downarrow$
&
&
\\

\midrule

$\checkmark$
&
& $\underline{26.746}$
& $\underline{0.8887}$
& $0.0892$
& $\underline{24.598}$
& $\underline{0.7130}$
& $\underline{0.3780}$
& $74.10$
& $\underline{56.38}$ \\

&
$\checkmark$
& $26.566$
& $0.8869$
& $\mathbf{0.0759}$
& $24.273$
& $0.6825$
& $0.4084$
& $\underline{74.13}$
& $55.65$ \\

$\checkmark$
& $\checkmark$
& $\mathbf{26.881}$
& $\mathbf{0.8943}$
& $\underline{0.0842}$
& $\mathbf{25.045}$
& $\mathbf{0.7616}$
& $\mathbf{0.3460}$
& $\mathbf{74.26}$
& $\mathbf{57.32}$ \\

\bottomrule
\end{tabular}%
}
\end{table}



\section{Conclusion}
\label{ch:conclusion}

We proposed DAP-MR, a factorized conditioning framework for ultrasound super-resolution that separates degradation identity from magnitude. CLIP-initialized degradation tokens are adapted during SR training, while an input-dependent router provides continuous magnitudes for conditioning. Partial-dose and mismatch-specificity supervision encourage magnitude-aware and degradation-selective restoration. Experiments across one-step and multi-step diffusion SR backbones show stronger gains under anatomical-domain shift and on unpaired real-world ultrasound data. Magnitude sweeps further confirm sensitivity to the injected control. Overall, DAP-MR provides a flexible degradation-aware conditioning strategy for ultrasound SR.

\begin{acks}
This work was financially supported in part by the Co-creation Platform of the Industry Academia Innovation School, NYCU (project number: 112UA10019), under the framework of the National Key Fields Industry-University Cooperation and Skilled Personnel Training Act, from the Ministry of Education (MOE) and industry partners in Taiwan. It also supported in part by the National Science and Technology Council, Taiwan, under Grant NSTC-115-2634-F-A49-011-, NSTC-114-2218-E-A49-024-, Grant NSTC-115-2425-H-A49-001, Grant NSTC-114-2622-E-A49-027, Grant NSTC-115-2221-E-A49 -124 -MY3, Grant NSTC-115-2218-E-A49 -017 and in part by the Higher Education Sprout Project of the National Yang Ming Chiao Tung University and the Ministry of Education (MOE), Taiwan. It is also partly supported by MediaTek Inc., Hon Hai Research Institute, E.SUN Bank, and Industrial Technology Research Institute. (Corresponding author: Ching-Chun Huang.)
\end{acks}

\FloatBarrier
\bibliographystyle{ACM-Reference-Format}
\bibliography{sample-base}

@String(IJCV  = {Int. J. Comput. Vis.})

@String(CVPR  = {IEEE Conf. Comput. Vis. Pattern Recog.})

@String(ICCV  = {Int. Conf. Comput. Vis.})

@String(ECCV  = {Eur. Conf. Comput. Vis.})

@String(IJCV  = {IJCV})

@String(CVPR  = {CVPR})

@String(ICCV  = {ICCV})

@String(ECCV  = {ECCV})

@article{camus,
  title={Deep Learning for Segmentation Using an Open Large-Scale Dataset in 2D Echocardiography},
  author={Leclerc, Sarah and Smistad, Erik and Pedrosa, Jo{\~a}o and Ostvik, Andreas and Cervenansky, Frederik and Espinosa, Fabrice and Espeland, Torvald and Berg, Erik Andreas Rye and Jodoin, Pierre-Marc and Grenier, Thomas and Lartizien, Carole and Dhooge, Jan and Lovstakken, Lasse and Bernard, Olivier},
  journal={IEEE Transactions on Medical Imaging},
  volume={38},
  number={9},
  pages={2198--2210},
  year={2019},
  doi={10.1109/TMI.2019.2900516}
}

@article{busi,
  title={Dataset of Breast Ultrasound Images},
  author={Al-Dhabyani, Walid and Gomaa, Mohammed and Khaled, Hussien and Fahmy, Aly},
  journal={Data in Brief},
  volume={28},
  pages={104863},
  year={2020},
  doi={10.1016/j.dib.2019.104863}
}

@inproceedings{realesrgan,
  title={Real-esrgan: Training real-world blind super-resolution with pure synthetic data},
  author={Wang, Xintao and Xie, Liangbin and Dong, Chao and Shan, Ying},
  booktitle={Proceedings of the IEEE/CVF international conference on computer vision},
  pages={1905--1914},
  year={2021}
}

@inproceedings{seesr,
  title={Seesr: Towards semantics-aware real-world image super-resolution},
  author={Wu, Rongyuan and Yang, Tao and Sun, Lingchen and Zhang, Zhengqiang and Li, Shuai and Zhang, Lei},
  booktitle=CVPR,
  pages={25456--25467},
  year={2024}
}

@inproceedings{pisasr,
  title={Pixel-level and semantic-level adjustable super-resolution: A dual-lora approach},
  author={Sun, Lingchen and Wu, Rongyuan and Ma, Zhiyuan and Liu, Shuaizheng and Yi, Qiaosi and Zhang, Lei},
  booktitle=CVPR,
  pages={2333--2343},
  year={2025}
}

@article{stablesr,
  title={Exploiting diffusion prior for real-world image super-resolution},
  author={Wang, Jianyi and Yue, Zongsheng and Zhou, Shangchen and Chan, Kelvin CK and Loy, Chen Change},
  journal=IJCV,
  volume={132},
  number={12},
  pages={5929--5949},
  year={2024},
  publisher={Springer}
}

@article{promptsr,
  title={Image super-resolution with text prompt diffusion},
  author={Chen, Zheng and Zhang, Yulun and Gu, Jinjin and Yuan, Xin and Kong, Linghe and Chen, Guihai and Yang, Xiaokang},
  journal={arXiv preprint arXiv:2311.14282},
  year={2023}
}

@inproceedings{tvt,
  title={Fine-structure preserved real-world image super-resolution via transfer vae training},
  author={Yi, Qiaosi and Li, Shuai and Wu, Rongyuan and Sun, Lingchen and Wu, Yuhui and Zhang, Lei},
  booktitle=ICCV,
  pages={12415--12426},
  year={2025}
}

@article{s3diff,
  title={Degradation-guided one-step image super-resolution with diffusion priors},
  author={Zhang, Aiping and Yue, Zongsheng and Pei, Renjing and Ren, Wenqi and Cao, Xiaochun},
  journal={arXiv preprint arXiv:2409.17058},
  year={2024}
}

@article{dalpsr,
  title={DaLPSR: Leverage degradation-aligned language prompt for real-world image super-resolution},
  author={Jiang, Aiwen and Wei, Zhi and Peng, Long and Liu, Feiqiang and Wang, Mingwen},
  journal={Image and Vision Computing},
  volume={170},
  pages={105978},
  year={2026},
  publisher={Elsevier}
}

@inproceedings{cyclegan,
  title={Unpaired image-to-image translation using cycle-consistent adversarial networks},
  author={Zhu, Jun-Yan and Park, Taesung and Isola, Phillip and Efros, Alexei A},
  booktitle=ICCV,
  pages={2223--2232},
  year={2017}
}

@article{prlmu,
  title={Progressive residual learning with memory upgrade for ultrasound image blind super-resolution},
  author={Liu, Heng and Liu, Jianyong and Chen, Feng and Shan, Caifeng},
  journal={IEEE Journal of Biomedical and Health Informatics},
  volume={26},
  number={9},
  pages={4390--4401},
  year={2022},
  publisher={IEEE}
}

@article{cardiac,
  title={Progressive residual learning with memory upgrade for ultrasound image blind super-resolution},
  author={Liu, Heng and Liu, Jianyong and Chen, Feng and Shan, Caifeng},
  journal={IEEE Journal of Biomedical and Health Informatics},
  volume={26},
  number={9},
  pages={4390--4401},
  year={2022},
  publisher={IEEE}
}

@misc{stanford2020echonet,
  author    = {Stanford AIMI Center},
  title     = {EchoNet-Dynamic Cardiac Ultrasound},
  year      = {2020},
  publisher = {Stanford Center for Artificial Intelligence in Medicine and Imaging},
  doi       = {10.71718/yqp5-y078}
}

@article{HMC_QU,
  author={Degerli, Aysen and Zabihi, Morteza and Kiranyaz, Serkan and Hamid, Tahir and Mazhar, Rashid and Hamila, Ridha and Gabbouj, Moncef},
  journal={IEEE Access}, 
  title={Early Detection of Myocardial Infarction in Low-Quality Echocardiography}, 
  year={2021},
  volume={9},
  number={},
  pages={34442-34453},
  doi={10.1109/ACCESS.2021.3059595}
}

@inproceedings{SAMUS,
  title={Beyond adapting SAM: Towards end-to-end ultrasound image segmentation via auto prompting},
  author={Lin, Xian and Xiang, Yangyang and Yu, Li and Yan, Zengqiang},
  booktitle={International Conference on Medical Image Computing and Computer-Assisted Intervention},
  pages={24--34},
  year={2024},
  organization={Springer}
}

@inproceedings{spire,
  title={Spire: Semantic prompt-driven image restoration},
  author={Qi, Chenyang and Tu, Zhengzhong and Ye, Keren and Delbracio, Mauricio and Milanfar, Peyman and Chen, Qifeng and Talebi, Hossein},
  booktitle=ECCV,
  pages={446--464},
  year={2024},
  organization={Springer}
}

@article{ul_sr25,
    title = {Blind super-resolution for handheld ultrasound image: Two-stage degradation based unpaired deep learning},
    journal = {Computer Methods and Programs in Biomedicine},
    volume = {270},
    pages = {108956},
    year = {2025},
    issn = {0169-2607},
    doi = {https://doi.org/10.1016/j.cmpb.2025.108956},
    author = {Zhencun Jiang and Kangrui Ren and Kefan Wang and Zhongjie Wang},
}

@article{ul_sr24,
    doi = {10.1088/1361-6560/ad4086},
    year = {2024},
    month = {jun},
    publisher = {IOP Publishing},
    volume = {69},
    number = {12},
    pages = {125026},
    author = {Liu, Tianyu and Han, Shuai and Xie, Linru and Xing, Wenyu and Liu, Chengcheng and Li, Boyi and Ta, Dean},
    title = {Super-resolution reconstruction of ultrasound image using a modified diffusion model},
    journal = {Physics in Medicine \& Biology},
}

@article{ul_sr26,
  title={Blind Ultrasound Image Enhancement via Self-Supervised Physics-Guided Degradation Modeling},
  author={Khan, Shujaat and Atif, Syed Muhammad and Huh, Jaeyoung and Azhar, Syed Saad},
  journal={arXiv preprint arXiv:2601.21856},
  year={2026}
}

@inproceedings{psnr,
  title={Image quality metrics: PSNR vs. SSIM},
  author={Hore, Alain and Ziou, Djemel},
  booktitle={2010 20th international conference on pattern recognition},
  pages={2366--2369},
  year={2010},
  organization={IEEE}
}

@inproceedings{lpips,
  title={The Unreasonable Effectiveness of Deep Features as a Perceptual Metric},
  author={Zhang, Richard and Isola, Phillip and Efros, Alexei A. and Shechtman, Eli and Wang, Oliver},
  booktitle=CVPR,
  pages={586--595},
  year={2018}
}

\clearpage
\appendix
\renewcommand\footnotetextcopyrightpermission[1]{} 

\definecolor{myorange}{RGB}{230,120,20}


\acmConference[MMAsia]{ACM Multimedia Asia 2026}{2026}{VinUniversity, Hanoi, Vietnam}

\begin{strip}
\centering

{\Large\bfseries
Supplementary Material for DAP-MR:
Degradation-Aware Prompt--Magnitude Routing for
Real-World Ultrasound Super-Resolution
\par}

\vspace{12pt}

\begin{minipage}[t]{0.44\textwidth}
    \centering
    \includegraphics[width=\linewidth]{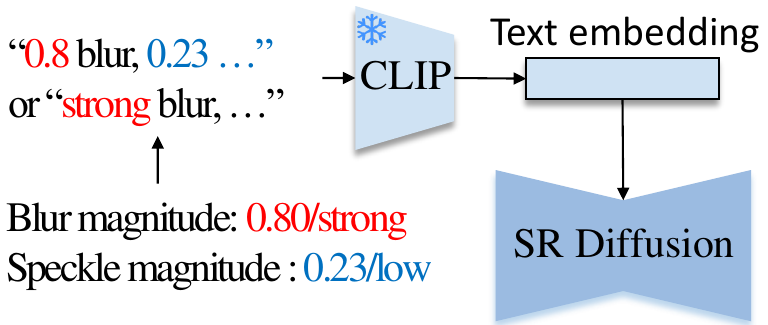}

    \small (a) Severity encoded in text
\end{minipage}
\hfill
\begin{minipage}[t]{0.54\textwidth}
    \centering
    \includegraphics[width=\linewidth]{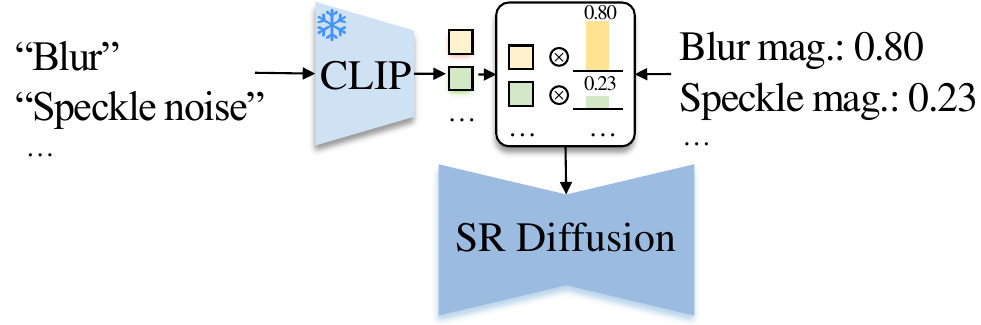}

    \small (b) Explicit magnitude routing
\end{minipage}

\vspace{3pt}

\begin{minipage}{0.96\textwidth}
    \centering
    \captionof{figure}{
        Comparison of degradation conditioning strategies.
        Existing methods encode degradation type and severity
        together in text, whereas DAP-MR retains degradation
        identity in text-derived tokens and controls restoration
        strength using explicit continuous magnitudes.
    }
    \label{fig:conditioning_comparison}
\end{minipage}

\end{strip}

\vspace{-4pt}

This supplementary material provides additional details omitted from the main manuscript due to space constraints. We provide the complete training procedures of DAP-MR, implementation details of the synthetic
degradation pipeline and magnitude router, and additional analyses and ablation studies.

\section{Overview}
Figure~\ref{fig:conditioning_comparison} summarizes the key distinction between conventional text-based degradation conditioning and DAP-MR. Existing approaches encode both degradation identity and severity in the text prompt, requiring the text encoder to represent \emph{what} degradation is present and \emph{how strong} it is within the same embedding space. In contrast, DAP-MR retains degradation identity in text-derived tokens while injecting the corresponding continuous magnitudes explicitly. This factorization provides a separate restoration-strength signal while preserving the original text-conditioning pathway of the SR backbone.

\section{Training and Implementation Details}
\label{sec:supp_training_algorithms}

This section describes the online degradation synthesis, magnitude-router
pretraining, and magnitude-aware training of the SR backbone.
Algorithm~\ref{alg:router_pretrain} summarizes router pretraining, while
Algorithm~\ref{alg:dapmr_training} presents the complete DAP-MR training
procedure.

\subsection{Online Degradation Synthesis}
\label{sec:supp_degradation}

We synthesize LR ultrasound images online from clean HR images using
$K=6$ degradation categories: blur, resize, depth-dependent attenuation,
speckle noise, electronic sensor noise, and anechoic haze. Each
degradation is associated with a scalar severity parameter $p_k$ and a
normalized magnitude
\begin{equation}
    m_k=\psi_k(p_k)\in[0,1].
\end{equation}
The parameterization is defined consistently across degradations such
that $m_k=0$ corresponds to no degradation and $m_k=1$ corresponds to
the maximum severity considered in our synthesis pipeline. Table~
\ref{tab:degradation_settings} summarizes the settings.

\begin{table*}[t]
\centering
\caption{Synthetic degradation settings used for online LR generation.
For each degradation, $p_k$ denotes its severity parameter and
$m_k=\psi_k(p_k)\in[0,1]$ is the normalized magnitude. The lower and
upper bounds correspond to no degradation and maximum severity,
respectively.}
\label{tab:degradation_settings}
\small
\setlength{\tabcolsep}{7pt}
\renewcommand{\arraystretch}{1.15}
\begin{tabular}{llll}
\toprule
\textbf{Degradation} &
\textbf{Severity parameter} &
\textbf{Sampling range} &
\textbf{Magnitude normalization} \\
\midrule

Blur &
Blur width $\sigma$ &
$\sigma\sim\mathcal{U}(0,3.0)$ &
$\displaystyle m_k=\frac{\sigma}{3.0}$ \\

Resize &
Downsampling scale $s$ &
$s\sim\mathcal{U}(1,4.0)$ &
$\displaystyle m_k=\frac{s-1}{3.0}$ \\

Attenuation &
Attenuation strength $\alpha_{\mathrm{att}}$ &
$\alpha_{\mathrm{att}}\sim\mathcal{U}(0,0.60)$ &
$\displaystyle m_k=\frac{\alpha_{\mathrm{att}}}{0.60}$ \\

Speckle &
Speckle strength $\alpha_{\mathrm{spk}}$ &
$\alpha_{\mathrm{spk}}\sim\mathcal{U}(0,0.35)$ &
$\displaystyle m_k=\frac{\alpha_{\mathrm{spk}}}{0.35}$ \\

Sensor noise &
Noise std. $\sigma_n$ &
$\sigma_n\sim\mathcal{U}(0,0.06)$ &
$\displaystyle m_k=\frac{\sigma_n}{0.06}$ \\

Anechoic haze &
Haze strength $h$ &
$h\sim\mathcal{U}(0,0.12)$ &
$\displaystyle m_k=\frac{h}{0.12}$ \\

\bottomrule
\end{tabular}
\end{table*}

The severity ranges are selected to cover a broad range of visually
plausible ultrasound degradations while avoiding excessively destructive
cases that remove substantial anatomical information. During online
synthesis, one or multiple degradation types may be applied to each
sample. A degradation that is not applied is assigned magnitude
$m_k=0$. A clean bypass is additionally used with probability $0.1$.

\subsection{Magnitude Router Pretraining}
\label{sec:supp_router_pretraining}

The magnitude router is pretrained using synthetically degraded
ultrasound images with known continuous magnitude labels. During this
stage, the VAE encoder is frozen and only the router parameters are
optimized. Given a synthetic LR image $\mathbf{x}_L$, the frozen VAE
encoder first produces a latent representation, from which the router
predicts the $K$ degradation magnitudes.

After pretraining, the router is frozen and used to provide
input-dependent magnitude estimates during SR training and blind
inference.

\begin{algorithm}[t]
\caption{Magnitude Router Pretraining}
\label{alg:router_pretrain}
\begin{algorithmic}[1]

\Require HR image dataset $\mathcal{D}_{H}$; degradation generator
$\mathcal{G}$; frozen VAE encoder $E$; magnitude router $R_{\phi}$;
total router iterations $N_R$

\For{$r=1$ to $N_R$}

    \State Sample $\mathbf{x}_{H}\sim\mathcal{D}_{H}$

    \State $(\mathbf{x}_{L},\mathbf{m})
    \gets \mathcal{G}(\mathbf{x}_{H})$
    \Comment{Online degradation synthesis}

    \State $\mathbf{z}\gets E(\mathbf{x}_{L})$

    \State $\hat{\mathbf{m}}
    \gets \sigma\!\left(R_{\phi}(\mathbf{z})\right)$

    \State $\mathcal{L}_{\mathrm{router}}
    \gets
    \dfrac{1}{K}
    \sum_{k=1}^{K}
    \operatorname{SmoothL1}
    \left(\hat{m}_{k},m_{k}\right)$

    \State Update $\phi$ using $\mathcal{L}_{\mathrm{router}}$

\EndFor

\Ensure Pretrained magnitude router $R_{\phi}$

\end{algorithmic}
\end{algorithm}

The router implementation and optimization settings are summarized in
Table~\ref{tab:router_settings}.

\begin{table}[t]
\centering
\caption{Magnitude-router implementation and training settings.}
\label{tab:router_settings}
\small
\begin{tabular}{lc}
\toprule
\textbf{Setting} & \textbf{Value} \\
\midrule
Input representation & Frozen VAE latent \\
Router architecture & MLP \\
Number of parameters & $7{,}089{,}030$ \\
Output dimension & $K=6$ \\
Output activation & Sigmoid \\
Training iterations $N_R$ & $3000$ \\
Optimizer & AdamW \\
Learning rate & $5\times10^{-5}$ \\
Batch size & $1$ \\
Weight decay & $0.01$ \\
Regression loss & Smooth-$L_1$ \\
Smooth-$L_1$ parameter & $\beta=1.0$ \\
\bottomrule
\end{tabular}
\end{table}

\subsection{Magnitude-Aware SR Training}
\label{sec:supp_sr_training}

After router pretraining, the router is fixed and the SR backbone is
optimized using three complementary training modes:
\textit{full restoration}, \textit{partial dose}, and
\textit{mismatch specificity}. Full restoration learns the clean
restoration endpoint, partial-dose supervision associates intermediate
magnitudes with intermediate restoration targets, and mismatch-specificity
supervision encourages degradation-selective responses.

For full-restoration supervision, ground-truth degradation magnitudes
are used during the initial stage of training. At iteration $aN$, the
conditioning source switches to router-predicted magnitudes. This
exposes the SR model to the estimated magnitude conditions used during
blind inference while retaining accurate supervision during early
optimization.

\begin{algorithm*}[t]
\caption{DAP-MR Magnitude-Aware SR Training}
\label{alg:dapmr_training}
\scriptsize
\begin{algorithmic}[1]

\Require HR image dataset $\mathcal{D}_{H}$; degradation generator
$\mathcal{G}$; fixed pretrained router $R_{\phi}$; frozen VAE encoder
$E$; frozen text encoder $\mathcal{T}$; degradation prompts
$\{P_k\}_{k=1}^{K}$; SR backbone $F_{\theta}$; total SR iterations $N$;
switching ratio $a$

\For{$k=1$ to $K$}
    \State $\mathbf{t}_{k}
    \gets \mathcal{T}(P_k)_{\mathrm{EOS}}$
    \Comment{Initialize degradation token}
\EndFor

\For{$n=1$ to $N$}

    \State Sample $\mathbf{x}_{H}\sim\mathcal{D}_{H}$

    \State Sample
    $q\in\{\textit{full},\textit{partial},\textit{mismatch}\}$

    \If{$q=\textit{full}$}

        \State $(\mathbf{x}_{L}^{(q)},\mathbf{m})
        \gets \mathcal{G}(\mathbf{x}_{H})$
        \Comment{Single or multiple degradations}

        \State $\mathbf{y}^{(q)}\gets\mathbf{x}_{H}$

        \If{$n<aN$}
            \State $\tilde{\mathbf{m}}^{(q)}\gets\mathbf{m}$
        \Else
            \State $\mathbf{z}\gets E(\mathbf{x}_{L}^{(q)})$

            \State $\hat{\mathbf{m}}
            \gets\sigma\!\left(R_{\phi}(\mathbf{z})\right)$

            \State $\tilde{\mathbf{m}}^{(q)}
            \gets\hat{\mathbf{m}}$
        \EndIf

    \ElsIf{$q=\textit{partial}$}

        \State Sample degradation type $k$, magnitude $m_k$, and
        $u\in(0,1)$

        \State $\mathbf{x}_{L}^{(q)}
        \gets\mathcal{G}_{k}(\mathbf{x}_{H};m_k)$

        \State $\tilde{\mathbf{m}}^{(q)}\gets\mathbf{0}$

        \State $\tilde{m}_{k}^{(q)}\gets u m_k$

        \State $\mathbf{y}^{(q)}
        \gets
        \mathcal{G}_{k}
        \left(
            \mathbf{x}_{H};
            (1-u)m_k
        \right)$

    \Else
        \Comment{$q=\textit{mismatch}$}

        \State Sample degradation type $k$, magnitude $m_k$, and
        non-matching token $j\neq k$

        \State $\mathbf{x}_{L}^{(q)}
        \gets\mathcal{G}_{k}(\mathbf{x}_{H};m_k)$

        \State $\tilde{\mathbf{m}}^{(q)}\gets\mathbf{0}$

        \State $\tilde{m}_{j}^{(q)}\gets m_k$

        \State $\mathbf{y}^{(q)}\gets\mathbf{x}_{L}^{(q)}$

    \EndIf

    \For{$k=1$ to $K$}
        \State $\mathbf{c}_{k}^{(q)}
        \gets\tilde{m}_{k}^{(q)}\mathbf{t}_{k}$
    \EndFor

    \State $\mathbf{P}^{(q)}
    \gets[\mathbf{c}_{1}^{(q)},\ldots,\mathbf{c}_{K}^{(q)}]$

    \State $\hat{\mathbf{o}}^{(q)}
    \gets
    F_{\theta}
    \left(
        \mathbf{x}_{L}^{(q)},
        \mathbf{P}^{(q)}
    \right)$
    \Comment{Backbone-specific prediction}

    \State $\mathcal{L}_{\mathrm{SR}}^{(q)}
    \gets
    \mathcal{L}_{\mathrm{base}}
    \left(
        \hat{\mathbf{o}}^{(q)},
        \mathbf{y}^{(q)}
    \right)$

    \State Update $\theta$ using $\mathcal{L}_{\mathrm{SR}}^{(q)}$

\EndFor

\Ensure Trained magnitude-aware SR model $F_{\theta}$

\end{algorithmic}
\end{algorithm*}

Here, $\hat{\mathbf{o}}^{(q)}$ denotes the backbone-specific prediction.
Depending on the SR backbone, it may correspond to a restored image,
latent representation, or diffusion denoising prediction.
$\mathcal{L}_{\mathrm{base}}$ retains the original backbone-specific
loss terms, weights, and optimization settings. DAP-MR therefore
modifies the conditioning representation and supervision targets without
introducing an additional restoration loss.

The magnitude values in DAP-MR should be interpreted as continuous
conditioning variables rather than exact percentages of degradation
removal. A magnitude of zero corresponds to the no-degradation endpoint
of the corresponding synthetic operator and disables its
degradation-specific conditioning token. However, it does not disable
the underlying visual restoration pathway of the SR backbone. Similarly,
a magnitude of one denotes the maximum degradation-conditioned response
considered during training rather than a guarantee of exact HR recovery.

\begin{figure}[t]
    \centering
    \includegraphics[width=0.45\textwidth]{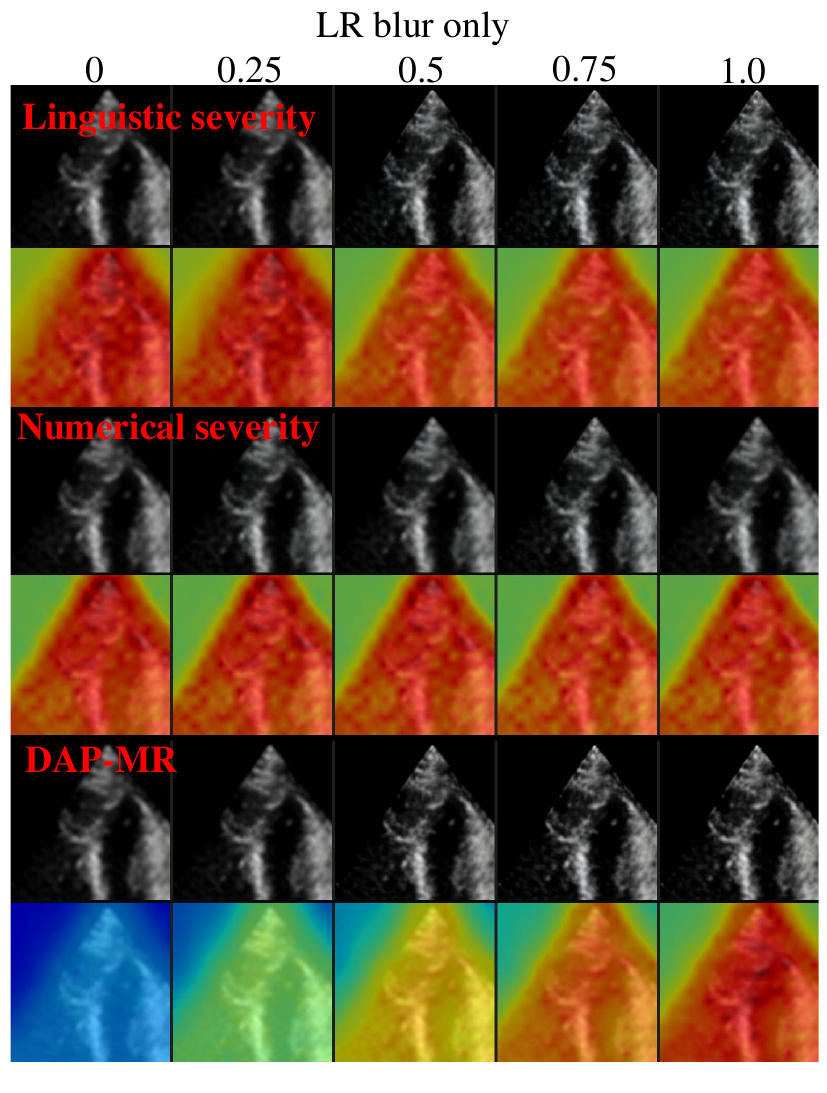} 
    \caption{\textbf{Magnitude-control comparison under blur degradation.}
    Given the same blur-degraded LR input, we vary the requested restoration strength
    from $0$ to $1$ using linguistic severity, numerical severity, and the proposed
    DAP-MR. DAP-MR exhibits the clearest and most progressive response to increasing
    magnitude, producing increasingly sharper structural details as the control value
    approaches $1$, whereas text-based severity representations show weaker or less
    consistent changes.
    }
    \label{fig:magnitude_control}
\end{figure}

\begin{figure*}[t]
    \centering
    \begin{subfigure}[t]{0.48\textwidth}
        \centering
        \includegraphics[width=\linewidth]{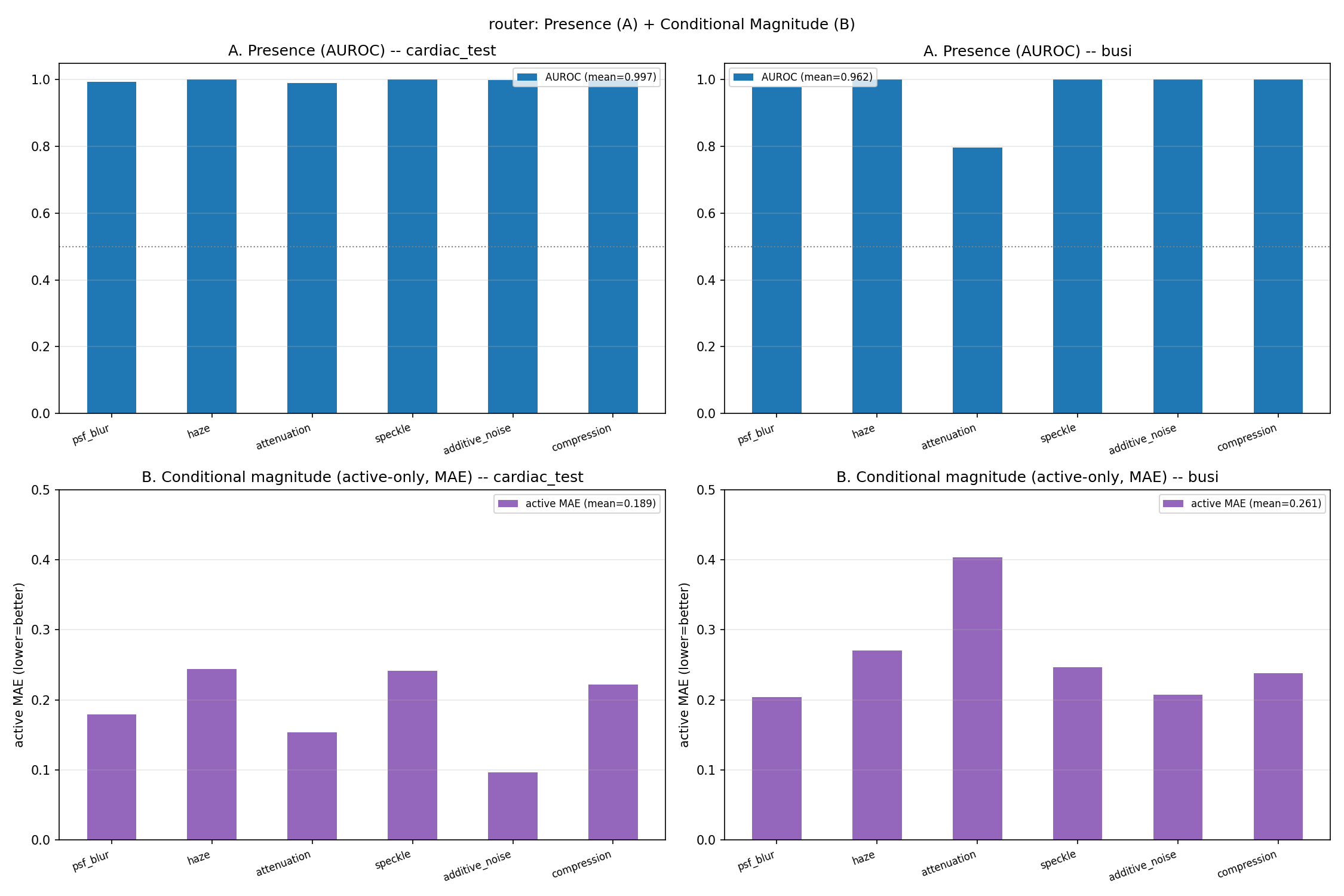}
        \caption{Degradation presence detection and conditional magnitude estimation.}
        \label{fig:router_ab}
    \end{subfigure}
    \hfill
    \begin{subfigure}[t]{0.48\textwidth}
        \centering
        \includegraphics[width=\linewidth]{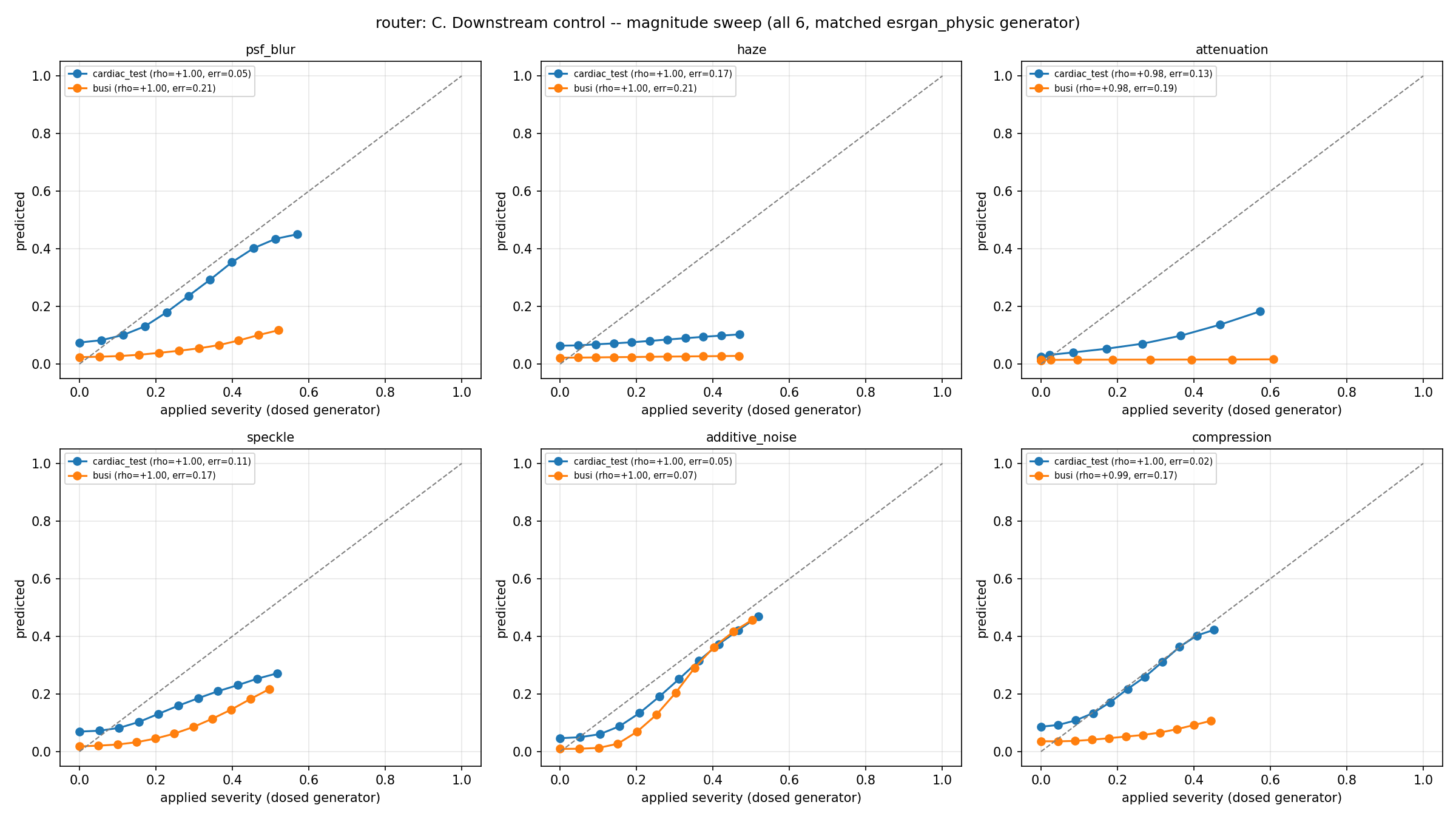}
        \caption{Router severity-response curves across six degradation types.}
        \label{fig:router_c}
    \end{subfigure}

    \caption{
    \textbf{Router evaluation on the in-domain Cardiac test set and the out-of-domain BUSI set.}
    Importantly, the router is \textbf{trained only on the in-domain Cardiac training set} and is evaluated on BUSI without target-domain adaptation.
    (a) AUROC measures whether the router correctly detects the presence of each degradation, while active-only MAE evaluates magnitude estimation when that degradation is present.
    (b) The severity-response curves show how the predicted magnitude changes as the applied degradation severity increases.
    The router reliably detects degradation presence and generally preserves severity ordering, although its magnitude estimates remain conservative, particularly on the out-of-domain BUSI set.
    }
    \label{fig:router_analysis}
\end{figure*}

\begin{table*}[t]
\centering
\caption{
Comparison of prompt representations using PiSA-SR.
Cardiac is in-domain, while BUSI, CAMUS-P, and HMC-QU are unseen during training.
DAP-MR uses degradation-specific EOS embeddings initialized from CLIP and jointly optimized during SR training.
Best and second-best results are shown in bold and underlined.
}
\label{tab:prompt_control_ablation}

\small
\setlength{\tabcolsep}{4pt}
\renewcommand{\arraystretch}{1.12}

\resizebox{\textwidth}{!}{%
\begin{tabular}{lcccccccc}
\toprule

\multirow{2}{*}{\textbf{Prompt strategy}}
&
\multicolumn{3}{c}{\textbf{Cardiac}}
&
\multicolumn{3}{c}{\textbf{BUSI}}
&
\multirow{2}{*}{\shortstack{\textbf{CAMUS-P}\\\textbf{Dice}$\uparrow$}}
&
\multirow{2}{*}{\shortstack{\textbf{HMC-QU}\\\textbf{Dice}$\uparrow$}} \\

\cmidrule(lr){2-4}
\cmidrule(lr){5-7}

&
\textbf{PSNR}$\uparrow$
&
\textbf{SSIM}$\uparrow$
&
\textbf{LPIPS}$\downarrow$
&
\textbf{PSNR}$\uparrow$
&
\textbf{SSIM}$\uparrow$
&
\textbf{LPIPS}$\downarrow$
&
&
\\

\midrule

Linguistic
& $26.800$
& $\underline{0.8931}$
& $\mathbf{0.0727}$
& $23.050$
& $0.6847$
& $0.4093$
& $74.11$
& $55.39$ \\

Numerical
& $26.867$
& $\mathbf{0.8943}$
& $0.0878$
& $24.906$
& $0.7469$
& $0.3674$
& $74.09$
& $54.89$ \\

RAMTag
& $\underline{26.877}$
& $0.8920$
& $0.1013$
& $\underline{25.031}$
& $\underline{0.7611}$
& $\underline{0.3590}$
& $\underline{74.24}$
& $\underline{56.40}$ \\

\textbf{DAP-MR (Fixed EOS)}
& $26.717$
& $0.8900$
& $\underline{0.0769}$
& $24.479$
& $0.7119$
& $0.3796$
& $74.11$
& $54.48$ \\

\textbf{DAP-MR (Learnable EOS)}
& $\mathbf{26.881}$
& $\mathbf{0.8943}$
& $0.0842$
& $\mathbf{25.045}$
& $\mathbf{0.7616}$
& $\mathbf{0.3460}$
& $\mathbf{74.26}$
& $\mathbf{57.32}$ \\

\bottomrule
\end{tabular}%
}
\end{table*}

\section{Additional Analysis}
\subsection{Magnitude-Control Analysis Across Degradation Types}
\label{sec:supp_controllability}

For each degradation, we synthesize a fixed LR input with severity $m=0.70$, bypass the router, and vary only the corresponding control magnitude as $s\in\{0,0.25,0.5,0.75,1.0\}$. We visualize the restored output $\hat{\mathbf{x}}(s)$, its residual $R_s=|\hat{\mathbf{x}}(s)-\mathbf{x}_H|$, and the signed residual change $\Delta R_s=R_0-R_s$, where positive values indicate improvement over zero control.

Across degradation types, controls around the true severity ($s=0.5$--$0.75$) generally produce the strongest positive effects, confirming that the injected magnitude meaningfully influences restoration. The response is not strictly monotonic, however, because text conditioning provides auxiliary guidance rather than fully controlling the U-Net; therefore, a milder control can occasionally outperform a closer or stronger magnitude. Blur is particularly sensitive to excessive control, which can introduce over-smoothing, consistent with the inherently smooth appearance of ultrasound structures. The response patterns are also degradation-specific: haze produces broad spatial changes, whereas sensor-noise control mainly affects scattered high-frequency regions. These distinct behaviors further indicate that the EOS degradation tokens encode degradation-specific information rather than acting as generic conditioning vectors. Note that $s=0$ disables only the corresponding degradation-specific condition, not the underlying SR backbone, while $s=1$ represents the maximum conditioning strength rather than guaranteed exact HR recovery.

\begin{table}[t]
    \centering
    \caption{Parameter and computational overhead of DAP-MR on PiSA-SR at $512\times512$ resolution. FLOPs include profiled convolution and matrix multiplication operations.}
    \label{tab:computational_overhead}
    \scriptsize
    \setlength{\tabcolsep}{4pt}
    \renewcommand{\arraystretch}{1.1}
    \begin{tabular}{llrr}
        \toprule
        \textbf{Backbone} &
        \textbf{Configuration} &
        \textbf{Params (M)} &
        \textbf{FLOPs (G)} \\
        \midrule
        \multirow{3}{*}{PiSA-SR}
        & Without router & 1289.966 & 4338.049 \\
        & With router    & 1297.055 & 4338.341 \\
        & \textbf{Overhead} &
        \textbf{+7.089} & \textbf{+0.292} \\
        \bottomrule
    \end{tabular}
\end{table}

\subsection{Magnitude Sensitivity Across Degradation Types}

We investigate the sensitivity of DAP-MR to continuous magnitude conditioning across six degradation types: blur, resolution degradation, depth-dependent attenuation, speckle noise, electronic sensor noise, and anechoic haze. For each type, we synthesize a fixed LR input at severity $m=0.70$, bypass the router, and manually vary the corresponding control magnitude $s\in\{0,0.25,0.5,0.75,1.0\}$ while keeping the remaining controls at zero. We visualize the restored output $\hat{\mathbf{x}}(s)$, its residual $R_s=|\hat{\mathbf{x}}(s)-\mathbf{x}_H|$, and the signed residual change $\Delta R_s=R_0-R_s$, where positive values indicate improvement over zero control. Across degradation types, increasing the control magnitude produces distinct changes in restoration outputs and residual distributions. Blur and resolution controls mainly affect local structural details, whereas speckle and sensor-noise controls influence high-frequency regions, and attenuation and haze produce broader intensity changes. Intermediate magnitudes often reduce reconstruction residuals, although the response is not strictly monotonic and stronger conditioning does not necessarily improve restoration. These observations demonstrate degradation-specific magnitude sensitivity, while the control values represent conditioning strength rather than exact percentages of degradation removal.

\begin{figure*}[t]
    \centering
    \includegraphics[width=\textwidth]{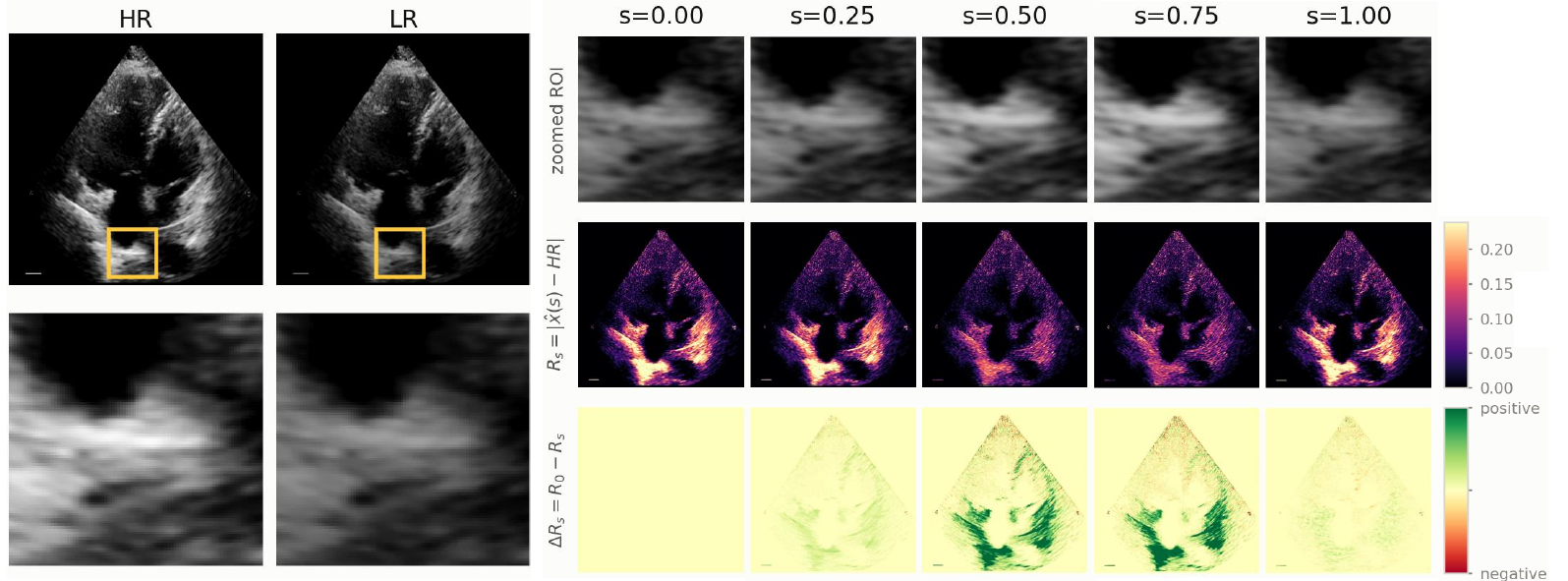}
    \caption{
    Magnitude sweep for depth-dependent attenuation at
    fixed LR degradation severity $m=0.70$.
    The router is bypassed, and the attenuation-control
    magnitude $s\in\{0,0.25,0.5,0.75,1.0\}$ is
    manually supplied. The rows show the restored
    output, zoomed ROI, residual $R_s$, and signed
    residual change $\Delta R_s=R_0-R_s$.
    }
    \label{fig:supp_sweep_attenuation}
\end{figure*}

\begin{figure*}[t]
    \centering
    \includegraphics[width=\textwidth]
    {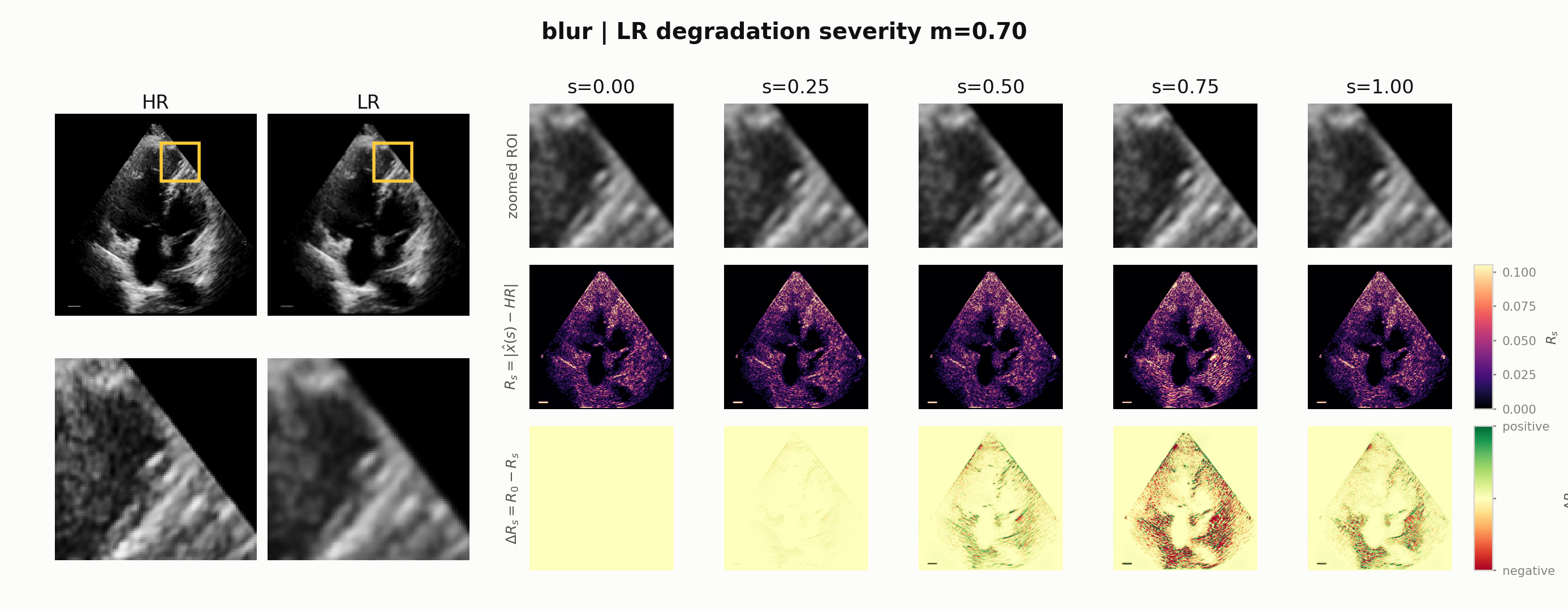}
    \caption{
    Magnitude sweep for blur at fixed LR degradation severity $m=0.70$.
    The router is bypassed and the blur-control magnitude
    $s\in\{0,0.25,0.5,0.75,1.0\}$ is manually supplied.
    The rows show the restored output, residual $R_s$, and signed
    residual change $\Delta R_s=R_0-R_s$.
    }
    \label{fig:supp_sweep_blur}
\end{figure*}

\begin{figure*}[t]
    \centering
    \includegraphics[width=\textwidth]
    {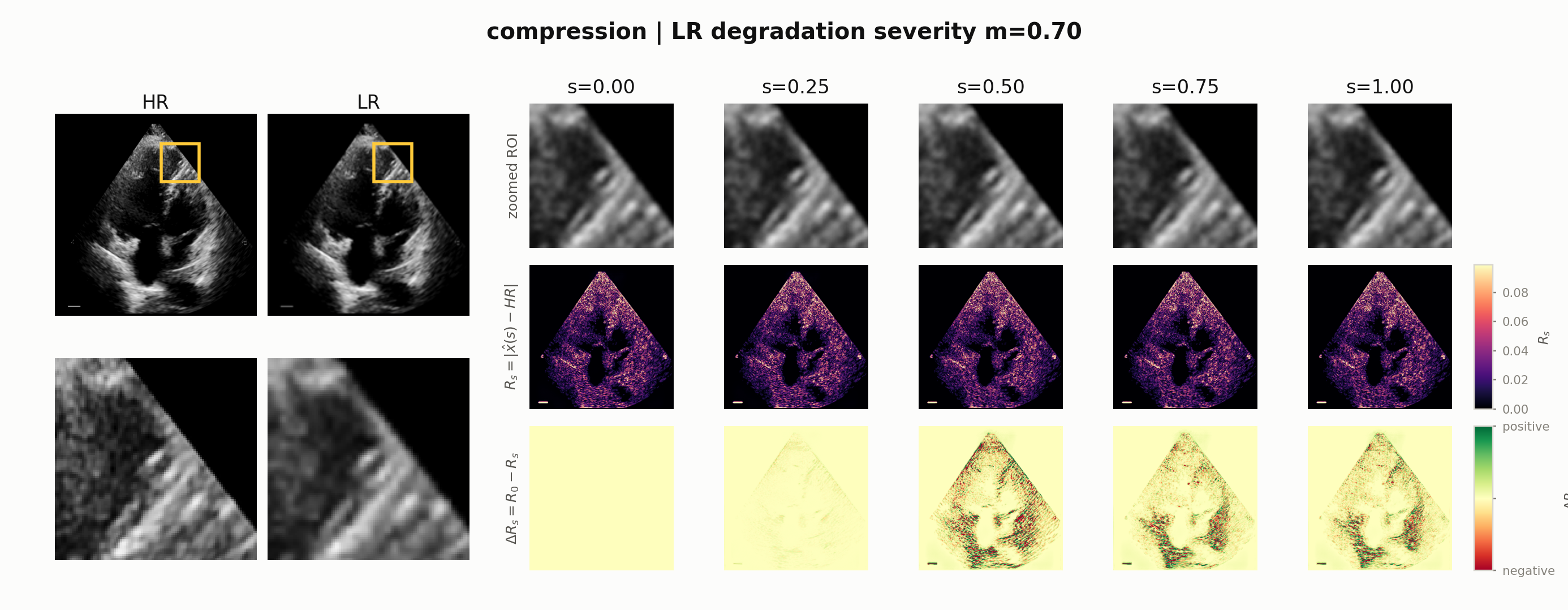}
    \caption{
    Magnitude sweep for resolution degradation at fixed LR degradation
    severity $m=0.70$. The router is bypassed and the corresponding
    control magnitude is varied from $s=0$ to $s=1$.
    Positive regions in $\Delta R_s$ indicate reduced reconstruction
    residual relative to the zero-control baseline.
    }
    \label{fig:supp_sweep_resolution}
\end{figure*}

\begin{figure*}[t]
    \centering
    \includegraphics[width=\textwidth]
    {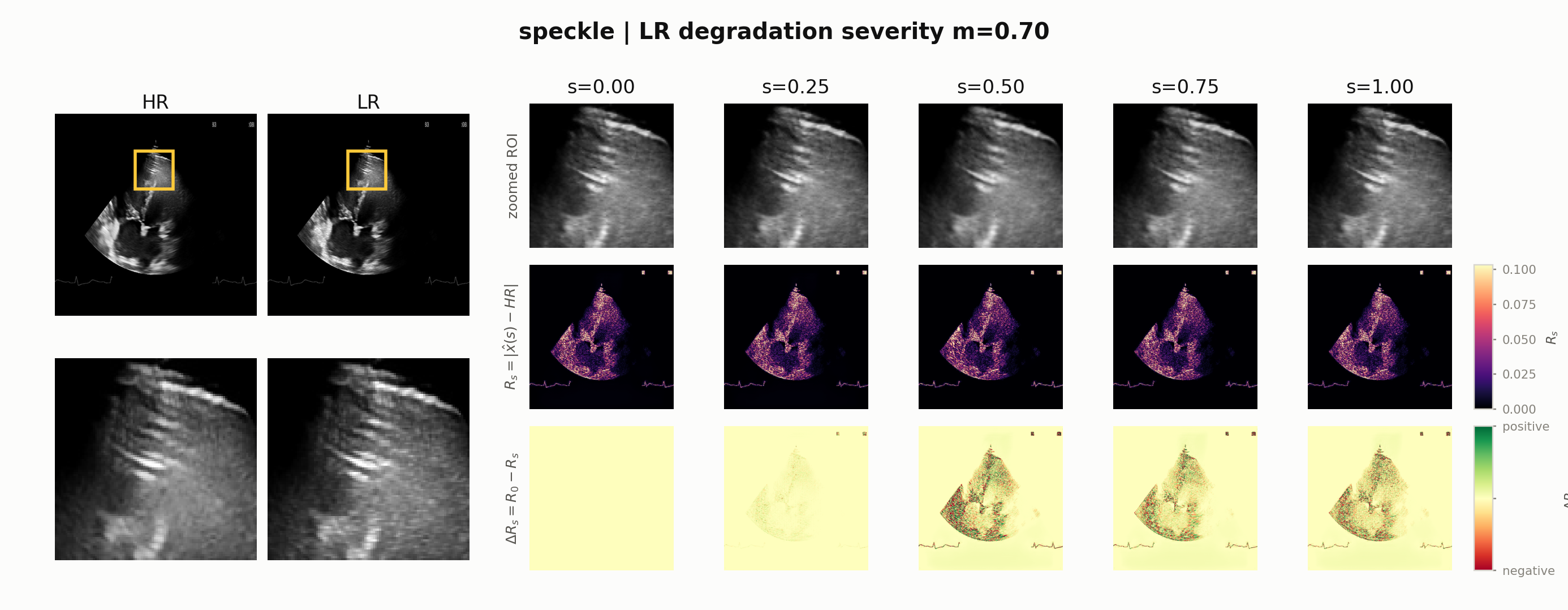}
    \caption{
    Magnitude sweep for speckle noise at fixed LR degradation severity
    $m=0.70$. Increasing $s$ changes only the speckle-specific
    conditioning magnitude while the remaining degradation controls are
    kept at zero.
    }
    \label{fig:supp_sweep_speckle}
\end{figure*}

\begin{figure*}[t]
    \centering
    \includegraphics[width=\textwidth]
    {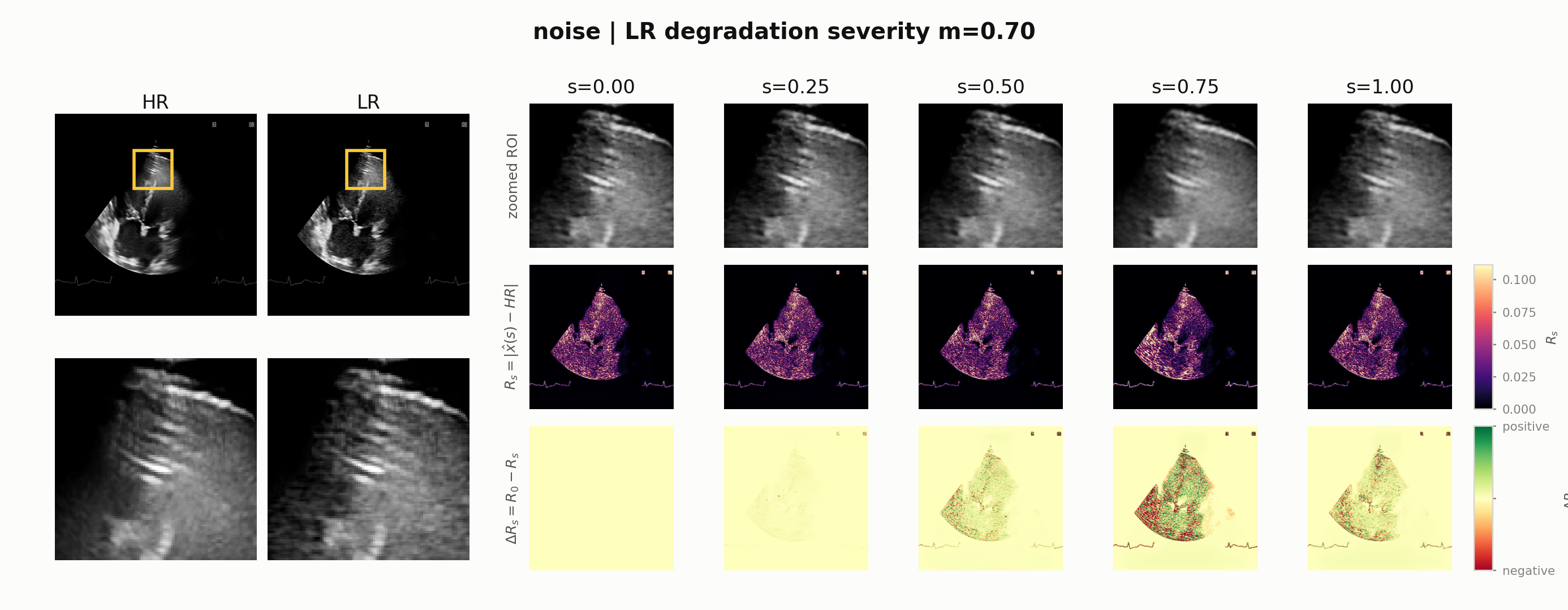}
    \caption{
    Magnitude sweep for electronic sensor noise at fixed LR degradation
    severity $m=0.70$. The output and residual maps illustrate how the
    restoration response changes as the noise-specific conditioning
    magnitude is increased.
    }
    \label{fig:supp_sweep_noise}
\end{figure*}

\begin{figure*}[t]
    \centering
    \includegraphics[width=\textwidth]
    {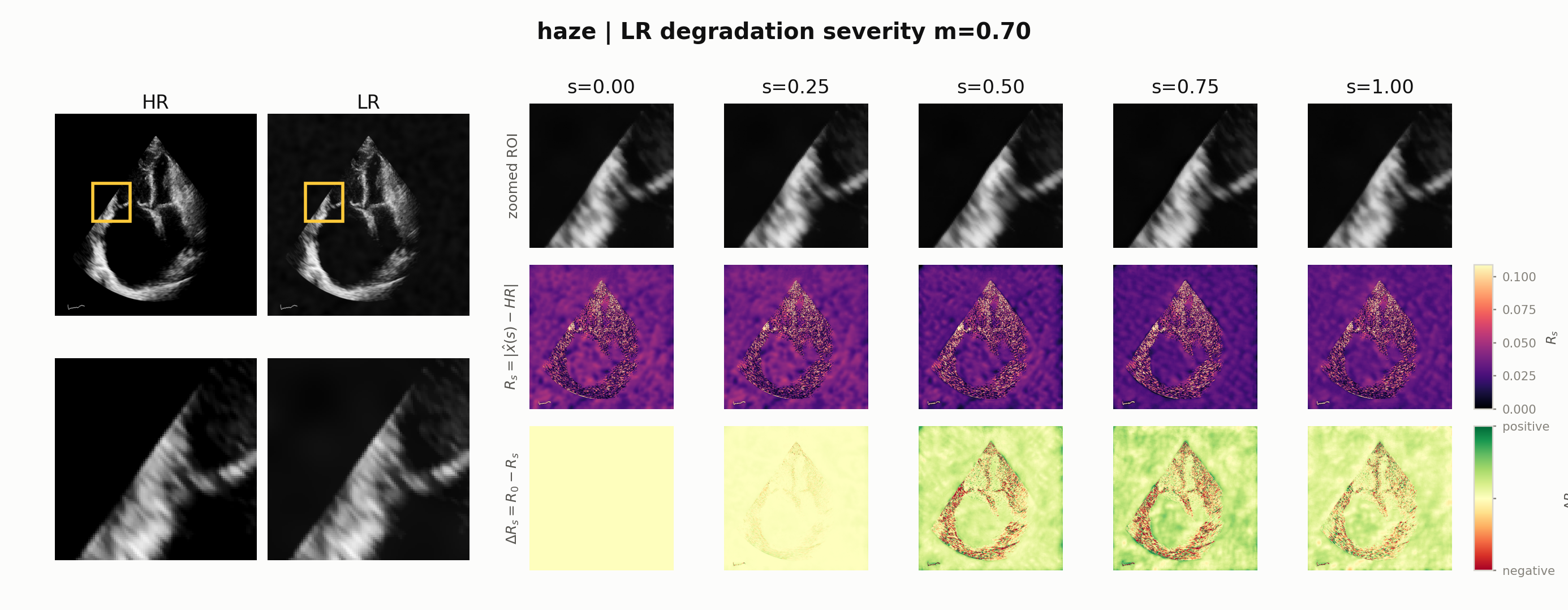}
    \caption{
    Magnitude sweep for anechoic haze at fixed LR degradation severity
    $m=0.70$. The router is bypassed and the haze-control magnitude is
    manually varied over $s\in\{0,0.25,0.5,0.75,1.0\}$.
    }
    \label{fig:supp_sweep_haze}
\end{figure*}

\subsection{Magnitude Controllability.} 
Figure~\ref{fig:magnitude_control} compares how different severity representations affect restoration under a fixed blur-degraded LR input. Linguistic severity produces only small changes across discrete levels, while numerical severity provides finer control values but still yields relatively weak and inconsistent output variations. In contrast, DAP-MR produces a substantially stronger and more progressive response as the supplied magnitude increases, showing that explicitly modulating the degradation token is more effective than relying on the text encoder to interpret severity from language alone. Importantly, the response should not be interpreted as an exact restoration percentage: stronger control does not always imply better reconstruction, since excessive blur correction may over-smooth ultrasound structures. Together with the degradation sweeps in Sec.~\ref{sec:supp_controllability}, these results support our design choice of separating degradation identity from magnitude: the text-derived token specifies \emph{what} degradation should be addressed, while the continuous magnitude provides an explicit signal for \emph{how strongly} it should influence restoration.

\subsection{Router Analysis.}
Figure~\ref{fig:router_analysis} evaluates the router from two complementary perspectives: degradation presence and magnitude estimation. Importantly, the router is \textbf{trained exclusively on synthetic degradations generated from the in-domain Cardiac training set}, while BUSI is never used during router training. Despite this anatomical-domain shift, the router maintains high AUROC on BUSI, indicating that degradation presence can be recognized without target-domain adaptation. Magnitude estimation is more challenging: the active-only MAE increases on BUSI, particularly for attenuation. Nevertheless, the severity-response curves generally preserve the correct ordering, with stronger applied degradations producing larger predicted magnitudes across most degradation types. These results suggest that the router transfers degradation identity and relative severity reasonably well beyond its training anatomy, although absolute magnitude calibration remains sensitive to domain shift. In particular, predictions become more conservative on BUSI and tend to underestimate stronger degradations. Since DAP-MR uses the router to provide an input-adaptive conditioning strength rather than an exact physical degradation estimate, preserving degradation presence and relative severity ordering is especially important. The remaining calibration gap also motivates the ground-truth-to-router transition during SR training, which exposes the backbone to imperfect router predictions before blind inference.

\subsection{Prompt Strategies}
Table~\ref{tab:prompt_control_ablation} compares alternative prompt representations under the same PiSA-SR setting. Linguistic prompting achieves the best Cardiac LPIPS ($0.0727$) but generalizes poorly to BUSI, where PSNR drops to $23.050$ dB and LPIPS increases to $0.4093$. In contrast, making the CLIP-initialized degradation EOS embeddings learnable provides the strongest overall cross-domain performance. Compared with fixed EOS tokens, the final DAP-MR improves BUSI PSNR from $24.479$ to $25.045$ dB and SSIM from $0.7119$ to $0.7616$, while reducing LPIPS from $0.3796$ to $0.3460$. It also improves HMC-QU Dice from $54.48$ to $57.32$. Although Cardiac LPIPS slightly increases from $0.0769$ to $0.0842$, the learnable EOS representation substantially improves transfer under anatomical-domain shift and is therefore adopted as the final DAP-MR configuration.


\end{document}